\documentclass[journal,compsoc]{IEEEtran}
\usepackage{amsmath,amssymb,amsfonts}
\usepackage{algorithmic}
\usepackage{graphicx}
\usepackage{textcomp}
\usepackage{xcolor}

\usepackage{graphicx}
\usepackage{subfig}
\usepackage{booktabs} 

\usepackage[ruled,vlined]{algorithm2e}
\usepackage{multirow}
\usepackage{makecell}
\usepackage{tikz}
\usetikzlibrary{positioning}
\usetikzlibrary{backgrounds}
\usepackage{tikzscale}
\usepackage{bm}
\usepackage{pgfplots}
\usepackage{pgfplotstable}
\usepackage{longtable}
\usepackage{url}
\usepgfplotslibrary{groupplots}
\pgfplotsset{compat=newest}
\usepackage{enumitem}
\usepackage{amsfonts}
\usepackage{amsmath}
\usepackage{mathtools} 
\usepackage{extarrows}
\usepackage{booktabs}
\usepackage{hyperref}
\usepackage{cleveref}
\DeclareUnicodeCharacter{220A}{\ensuremath{\in}}
\crefformat{section}{\S#2#1#3}
\crefformat{subsection}{\S#2#1#3}
\crefformat{subsubsection}{\S#2#1#3}
\crefformat{footnote}{#2\footnotemark[#1]#3}

\ifCLASSINFOpdf
\else
\fi

\begin{document}
%
\title{A Survey on Privacy in Graph Neural Networks:\\ Attacks, Preservation, and Applications}
%
%
%

\author{Yi Zhang\IEEEauthorrefmark{1}, \hspace{-0.2ex}Yuying Zhao\IEEEauthorrefmark{1}, \hspace{-0.2ex}Zhaoqing Li, \hspace{-0.2ex}Xueqi Cheng, \hspace{-0.2ex}Yu Wang, \hspace{-0.2ex}Olivera Kotevska, \hspace{-0.2ex}Philip S. Yu, \hspace{-0.2ex}and Tyler Derr

\thanks{\textsuperscript{*} denotes equal contribution and co-first authorship}
\IEEEcompsocitemizethanks{
\IEEEcompsocthanksitem Y. Zhang, Y. Zhao, X. Cheng, Y. Wang. and T. Derr are with Vanderbilt University, Nashville, TN, USA. Email: \{yi.zhang,yuying.zhao,xueqi.cheng,yu.wang.1,tyler.derr\}@vanderbilt.edu
\IEEEcompsocthanksitem Z. Li is with The Chinese University of Hong Kong, Hong Kong SAR, The People's Republic of China. Email: zhaoqingli@link.cuhk.edu.hk 
\IEEEcompsocthanksitem O. Kotevska is with the Oak Ridge National Laboratory, Oak Ridge, TN, USA. Email: kotevskao@ornl.gov
\IEEEcompsocthanksitem Philip S. Yu is with the University of Illinois Chicago, Chicago, IL, USA. Email: psyu@uic.edu
}
}

\markboth{} 
{Shell \MakeLowercase{\textit{et al.}}: Bare Demo of IEEEtran.cls for Computer Society Journals}
%



\IEEEtitleabstractindextext{%
\begin{abstract}
Graph Neural Networks (GNNs) have gained significant attention owing to their ability to handle graph-structured data and the improvement in practical applications. However, many of these models prioritize high utility performance, such as accuracy, with a lack of privacy consideration, which is a major concern in modern society where privacy attacks are rampant. To address this issue, researchers have started to develop privacy-preserving GNNs. Despite this progress, there is a lack of a comprehensive overview of the attacks and the techniques for preserving privacy in the graph domain. In this survey, we aim to address this gap by summarizing the attacks on graph data according to the targeted information, categorizing the privacy preservation techniques in GNNs, and reviewing the datasets and applications that could be used for analyzing/solving privacy issues in GNNs. We also outline potential directions for future research in order to build better privacy-preserving GNNs. 
\end{abstract}
\begin{IEEEkeywords}
Graph Neural Networks; Privacy Attacks; Privacy Preservation; Deep Learning on Graphs
\end{IEEEkeywords}
}

\maketitle

\IEEEdisplaynontitleabstractindextext

%
\IEEEpeerreviewmaketitle

\section{Introduction}

Graph-structured data, notable for its capacity to represent objects along with their interactions for a broad range of applications, is ubiquitous in the real world. Compared with independent and identically distributed (i.i.d) data that are typically utilized in deep neural networks (DNNs), graph data is more challenging to deal with due to its complexity in capturing object relationships and its irregular and non-grid-like shapes. To tackle the above challenges, various Graph Neural Networks (GNNs)~\cite{ma2021deep, wu2020comprehensive, rong2020deep, wu2022graph} 
are developed for multiple tasks such as node classification~\cite{Graphsage,kipf2016semi}, link prediction and recommendation~\cite{lu2011link,wu2022graph}, community detection~\cite{su2022comprehensive,shchur2019overlapping} and graph classification~\cite{li2019graph,wang2022imbalanced}. These models have achieved unprecedented success for applications across different domains such as e-commerce and recommender systems~\cite{wang2021graph,wu2022graph, wang2022cagcn}, social network analysis~\cite{perozzi2014deepwalk,derr2018signed}, financial quantitative analysis~\cite{wang2022review,sawhney2021stock}, and drug discovery~\cite{gilmer2017neural, liu2022interpretable}.

Despite their remarkable success in solving real-world tasks, most of GNNs lack privacy considerations. They are designed to achieve high performance, leaving private information vulnerable against attack.  Consequently, data privacy and safety from high-stake domains (e.g. finance, social, and medical) involving sensitive and private information could be undermined. In other words, without well-designed strategies, private information is constantly subjected to leakage. Even worse, a large variety of attack models are designed based on the vulnerability of the models. The aforementioned issues have become increasingly concerning, which spawned government regulations and laws for combating malicious attacks. For instance, the California Consumer Privacy Act (CCPA) was signed into law to protect customers’ privacy by regulating the collected information from businesses; the European Union has proposed a guideline that highlights the importance of trustworthy AI and indicates one of the ethical principles that a system should follow is the prevention of harm. Therefore, it is crucial to protect private data within GNNs (i.e., the privacy of the graph-structured data and the model parameters). 

However, the requirement for privacy protection in GNNs differs from that of the traditional DNNs. In addition to the need to protect sensitive features of node/graph instances, there is also the need to protect the relational information among entities in graphs which is at risk of exposure.
Furthermore, the unique message-passing mechanism exacerbates the challenge of protection since sensitive/confidential features might be potentially leaked during the propagation process. As a result, the existing privacy 
protecting methods developed for DNNs may not be readily adaptable to graphs, thereby imposing additional privacy requirements. These privacy requirements emerged uniquely in the graph domain have motivated a stream of work 
by cybersecurity experts and GNN researchers 
from academia and industry. 
In this work, we give a comprehensive survey about the attack strategies and privacy preservation techniques. We categorize the attack strategies into four categories, including Model Extraction Attacks (MEA)~\cite{wu2022model, defazio2019adversarial}, Graph Structure Reconstruction (GSR)~\cite{zhang2022model,he2021stealing,wu2022linkteller}, Attribute Inference Attacks (AIA)~\cite{duddu2020quantifying}, and Membership Inference Attacks (MIA)~\cite{olatunji2021membership,he2021node,duddu2020quantifying,hu2022membership}, where attackers aim to infer different parts of the graph data and GNN-based models. 
Specifically, MEA aims to extract a model that has similar behavior to the original model; GSR endeavors to reconstruct the graph structural information from limited information; AIA aims to infer the sensitive features, and MIA seeks to determine whether a certain component (e.g., node, edge, sub-graph) is contained in the training dataset. For privacy 
preserving techniques, they are summarized into four directions, namely, Latent Factor Disentangling~\cite{li2020adversarial, wanglearning, hu2022learning}, Adversarial Training~\cite{li2020adversarial, hsieh2021netfense, tian2021k}, Differentially Private Approach~\cite{sajadmanesh2020locally, sajadmanesh2022gap}, and Federated Learning~\cite{zhang2021federated, liu2022federated,liu2022federateds}. 
Generally, the goal of latent factor disentangling is to learn the representations that do not contain confidential information. Adversarial training aims to minimize the impact of specific attacks by reducing the performance of attacks during the training process. Differentially private approach utilizes differential privacy techniques to ensure data privacy. Federated learning~\cite{zhang2022robust}, on the other hand, seeks to develop distributed learning frameworks that enable various organizations to collaborate in training a model without sharing 
their own data.

This survey is primarily focused on investigating the privacy aspect of GNNs and the organization is as follows: 
We start by introducing the preliminary context and summarizing the relation to other surveys in Section~\ref{sec:context}, which includes the privacy concept on data, traditional privacy/attacks on deep learning, and the basic knowledge of graph data and deep learning on graphs.
We then present different types of privacy attack methods on GNNs in Section~\ref{sec:attack}. Thereafter, in Section~\ref{sec:preservation} we discuss privacy 
preserving techniques for GNNs. The currently used and more possible graph datasets for studying GNN privacy attacks/preservation along with applications are summarized in Section~\ref{sec: datasets}. After that, we discuss the future directions in Section~\ref{sec: future} and then conclude in Section~\ref{sec: conclusion}. 


\section{Preliminaries of Data Privacy, Attacks and Deep Learning on Graphs}
\label{sec:context}






In this section, we introduce the preliminaries for the privacy of data and models. We begin with the privacy of non-graph data and discuss the privacy and attacks in the general deep learning (DL) domain. After that, we introduce graph, graph data, and their deep learning techniques.
\subsection{Privacy on Data} \label{subsec: priv_tech}
Data privacy refers to protecting sensitive and confidential information such as personal identifiable information (PII) or information related to national security infrastructure. One standard approach to enhance the privacy of structured static data is to mask sensitive information, and it is used for query extraction, analysis, and sharing. Standard masking techniques such as $k$-anonymity, $l$-diversity, and $t$-closeness \cite{li2007t} are limited because another publicly accessible dataset
could be used to re-identify the masked entries. 

The above limitation allured a series of de-anonymization attacks aiming to extract or infer sensitive details or attributes associated with a specific record in the released dataset, such as isolation attacks and information amplification attacks 
To overcome this challenge, A group of researchers \cite{dwork2006differential} proposes privacy-preserving techniques called differential privacy (DP). The main idea is that if two datasets differ only by one record, then the same algorithm should have similar results on these datasets. It provides a solid mathematical privacy guarantee, which is formally defined to be the following:

\textbf{Definition 1:} \label{def:gdp} \textbf{Differential Privacy (DP)/Global Differential Privacy (GDP)}
A randomized mechanism $K$ gives $\epsilon$-DP if for all datasets $D$ and $D^\prime$ differing on at most one element, and all $S \subseteq Range(K)$,
\begin{equation*}
    Pr[K(D) \in S] \leq e^{\epsilon} Pr[K(D^\prime) \in S],
\end{equation*}
where $S$ is the set of all possible outputs of $K$; $K(D)$ is the privacy-preserving mechanism; and $Pr$ is the probability distribution of $K(D)$. The mechanism $K$ is guaranteed to leak information less than or equal to the amount specified by the parameter. The probability distributions of the randomized function $K(D)$ and $K(D^\prime)$ overlap, making it impossible to infer which of the two datasets $D$ and $D^\prime$ that the query is executed on. $\epsilon$ is a relative measurement of the allowance of information leakage and specifies how much the probability distributions should overlap~\cite{dwork2013toward}.

In DP, a data curator first collects the raw data and then performs an analysis (i.e., global differential privacy (GDP) defined in Def. \ref{def:ldp}). Local differential privacy (LDP) \cite{bebensee2019local} is a differential privacy in the local settings. In LDP, the data is perturbed first before being sent to an aggregator for analysis, which is the primary difference from GDP. 


\textbf{Definition 2:} \textbf{Local Differential Privacy (LDP)}\label{def:ldp}
A randomized mechanism $K$ satisfies $(\epsilon, \delta)$-LDP where $\epsilon \geq 0$ and $(0 \leq \delta \leq 1)$, if and only if any pair of input values $v, v' \in S$ and $S \subseteq Range(K)$, 
\begin{equation*}
    Pr[K(v) \in S] \leq e^{\epsilon} Pr[K(v') \in S],
\end{equation*}
where $Range(K)$ denotes the set of all possible outputs of the algorithm $K$. If $\delta = 0$, the algorithm  satisfies pure (strict) local differential privacy (pure LDP). If $\delta > 0$, the algorithm satisfies approximate (relaxed) local differential privacy (approximate LDP), namely, $(\epsilon, \delta)$-LDP \cite{xiong2020comprehensive}.

\begin{figure}[t]
    \centering
    \includegraphics[width = .99\columnwidth]{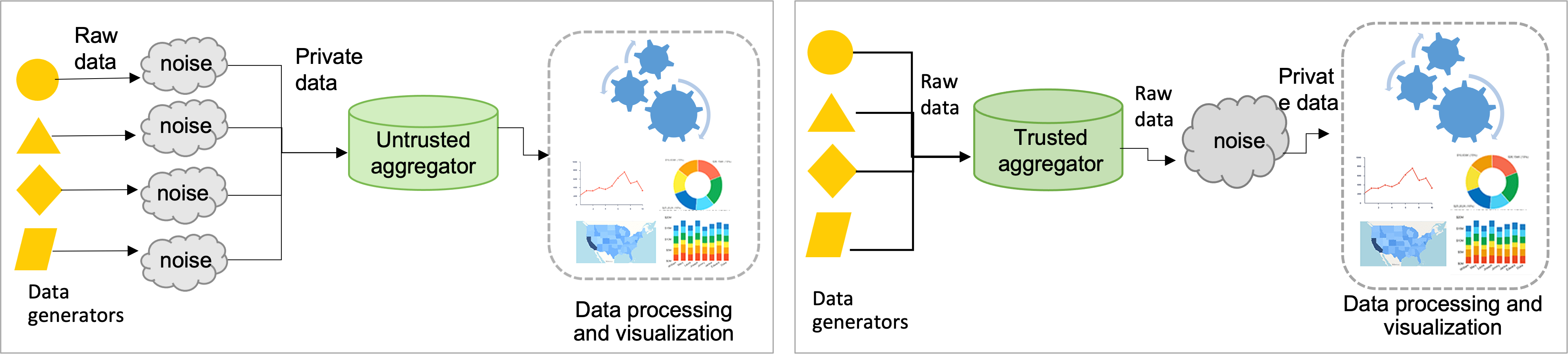}
    \caption{Visualizing the differences between local (left) and global (right) differential privacy.}
    \label{fig:global_vs_local}
\end{figure}

In summary, LDP is defined for the situation when individual data is privately protected before added to the database while GDP is defined for the situation when data is privately protected when it is queried from the database (see Figure \ref{fig:global_vs_local}, where we visualize LDP on the left and GDP is shown on the right). We note that both forms of DP are needed due to varying real-world application needs and settings.

\begin{figure*}[t]
    \centering
    \includegraphics[width = .90\textwidth]{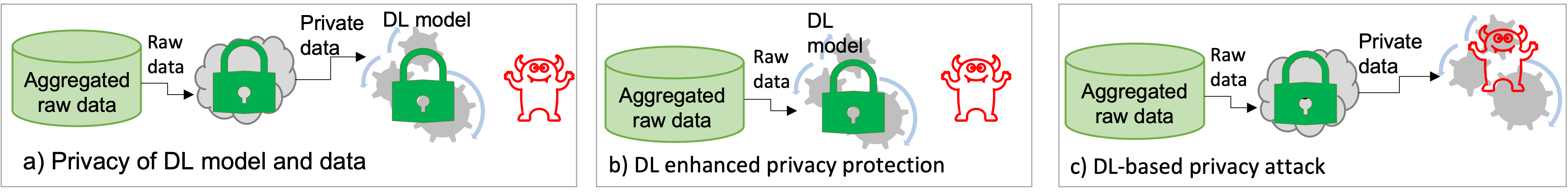}
    \vskip -1ex
    \caption{Categorization of research problems in privacy and deep learning.}
    \vspace{-1ex}
    \label{fig:ml_attacks}
\end{figure*}

In general, differential privacy can be achieved by adding a reasonable amount of random noise into the output results of the query function. The amount of noise will ultimately affect the trade-off between privacy and utility. Concretely, excessive noise will compromise the dataset while meager noise hampers privacy guarantees. Specifically, the amount of noise can be determined by sensitivity. Generally, there are two types of sensitivity, global sensitivity and local sensitivity, where we will next provide the associated definitions for both types of sensitivity as follows: 

\textbf{Definition 3: Global Sensitivity}
Given a query function $f$ that operates on a dataset $D$ and produces the maximal result difference for all datasets $(D, D^\prime)$ with at most one different entry, Global Sensitivity
is defined as:
\begin{equation*}
GS\left(f,D\right)={\max_{D,\ D^\prime}{||f(D)-f(D^\prime)||}_1}
\end{equation*} 
where $||.||_1$ is the $L_1$-norm distance between datasets differing at most one element, $max$ is the maximum result of $f(D)-f(D^\prime)$ for all datasets $D, D^\prime$.

\textbf{Definition 4: Local Sensitivity}
Given a query function $f$ that operates on a dataset $D$, the local sensitivity is the maximum difference that the change of one data point in $D$ can produce, defined as 
\begin{equation*}
LS\left(f,D\right)={\max_{D^\prime}{||f(D)-f(D^\prime)||}_1}
\end{equation*}
However, both GDP and LDP are highly vulnerable to manipulation (i.e., an adversary could insert additional data to undermine the output quality, i.e., poison attack). Additionally, most real-world data are unstructured, and the attacks and privacy challenges corresponding to such data will be discussed next.

\subsection{Privacy and Attacks on Deep Learning (DL)}
\label{sec.privacy_and_attacks_DL}

Next, we discuss the vulnerabilities of deep learning, present a categorization of privacy approaches with deep learning, and summarize the most common privacy attacks on deep learning. 

\subsubsection{Vulnerabilities of DL}

By leveraging a copious amount of data, deep learning (DL) algorithms are particularly impressive at decision-making, knowledge extraction, recommendation, forecasting, and many other crucial tasks. The input to DL models is the algebraic form (e.g. scalars, vectors, matrices, tensors, etc.) corresponding to raw images, videos, audios, text, graphs, and other data forms, and the output of DL models can be a class (classification), a value (regression), an embedding (encoding), or a generated sample (generative). Unfortunately, it is often possible to discern sensitive information from the input data based on the outputs of the neural network. During the training process, DL models encode the sensitive information of the training data, consequently, it is not surprising that a trained DL model could disclose sensitive information \cite{jagielski2020auditing}. The data is also vulnerable to attack because it is typically not obfuscated but instead stored in centralized repositories, which are subjected to the risks of data breaches. This type of data breach has been demonstrated in \cite{jagielski2020auditing}.
Under the context of private data analysis, we hope to ensure that anything that can be learned about a member of the database via a privacy-preserving database should also be learnable without access to the database \cite{dwork2013toward}. For example, in a medical database of smoking patients used for investigating if smoking causes cancer, the algorithm infers sensitive information from data even if the user's PII has been protected.


\subsubsection{Privacy Approaches for DL}
There are a few ways to protect the privacy of the DL model; Liu et al. \cite{liu2021machine} define them as follows (also seen in Figure~\ref{fig:ml_attacks}): 

\begin{enumerate}[label=(\alph*)]
\item \textbf{Privacy of DL model and data:} provides privacy protection to the DL model, training dataset, testing dataset, and the output because the assumption is that the whole DL system is the target for privacy protection.

\item \textbf{DL-enhanced privacy protection:} provides privacy protection of the data, and DL is a tool to help privacy protection. DL algorithm identifies the sensitive information and provides input to the user about privacy concerns.

\item \textbf{DL-based privacy attack:} DL is used as an attack tool by the adversary without access to the original dataset used for training. This is particularly important for cases when the model is trained to detect some sensitive information such as people’s identities and landmarks.
\end{enumerate}

However, there are vulnerabilities in private DL, and in the next part, we will explain the privacy attacks for each of the categories mentioned earlier. \par

\subsubsection{Privacy Attacks on DL}

Recent attacks against DL models \cite{rigaki2020survey,zhou2022adversarial} emphasize the implicit risks and catalyze an urgent demand for privacy preserving. Some of the work focused on efficient attack strategies while others focused on defense mechanism. The defense mechanism usually used differential privacy to provide guarantee. Differentially private DL ensures that the adversaries are incapable to infer any information about a single record with high confidence from the released DL models or output results \cite{al2019privacy}. However, in general, the attacks are split into white-box and black-box based on whether the model parameters of the target model are available. In \textit{White-box attack}, the attacker can access the target model. So it knows the gradients, architecture, hyper-parameters, and training data. In \textit{Black-box attack}, the attacker cannot access the target model but only has access to query the target model. Based on the interactions, the attackers then infer some information about the model, such as possible datasets used for training. 

There are many DL attack models proposed in the literature such as  model extraction attack, model inversion attack, attribute inference attack, and membership inference attack. We explain them briefly below. More details about the privacy attacks on graph neural networks are provided in Section \ref{sec:priv_gnn}.

\textbf{Model Extraction Attack (MEA):} The goal of this attack is to steal model parameters and hyper-parameters to duplicate or mimic the functionality of the target model. The adversary does not have any prior knowledge about the DL model parameters or training data. Wang et al. \cite{wang2018stealing} design an attack to get the hyper-parameters from the DL model. Tramer et al. \cite{tramer2016stealing} use the shadow model approach to get information about the target model. Takemura et al. \cite{takemura2020model} demonstrate the effectiveness of the MEA on complex neural networks such as recurrent neural networks with or without LSTM. Similarly, Zhang et al. \cite{zhang2021thief} demonstrate the MEA on pre-trained models.

\textbf{Model Inversion Attack (MIvA):} The goal of the model inversion attack is to use the output of the model to extract features that characterize one of the model's classes \cite{veale2018algorithms}. Fredrikson et al. \cite{fredrikson2015model} develop an attack that exploits confidence values revealed along with predictions. Hidano et al. \cite{hidano2017model} extend the work of \cite{fredrikson2015model} and assume no knowledge of the non-sensitive attributes. The focus of Parl et al. \cite{park2019attack} is on the defense side by using differential privacy. 

\textbf{Attribute Inference Attack (AIA):} AIA aims to reconstruct the missing attributes given partial information about the data record and access to the machine learning model. Jia et al. \cite{jia2018attriguard} develop a defense mechanism by adding noise to the sensitive attributes. Gong et al. \cite{gong2018attribute} develop a new attack to infer sensitive attributes of 
social network users. 

\textbf{Membership Inference Attack (MIA):} MIA aims to infer whether a data sample is part of the data used in training a model or not \cite{hu2022membership}. Shokri et al. \cite{shokri2017membership} use a shadow training technique to imitate the behavior of the target model. The trained model is used to discover 
differences in the target models on training and non-training inputs. Salem et al. \cite{salem2018ml} use an unsupervised binary classification instead of the shadow model. Truex et al. \cite{truex2019demystifying} study under what circumstances the model might be more vulnerable and find that collaborative learning exposes vulnerabilities to membership inference risks when the adversary is a participant. Jia et al. \cite{jia2019memguard} focus on defense mechanisms by adding noise to each confidence score vector predicted by the target classifier.

\subsection{Graphs and Graph-structured Data}
\label{sec:gnn-notation}
Graphs are powerful for representing relational data, and they are widely applied in different fields such as recommender system~\cite{wang2022cagcn, wu2022graph}, chemistry~\cite{liu2022interpretable, gilmer2017neural}, social science~\cite{wang2022improving, derr2018signed}, and e.t.c. Here we provide the graph notations that will be used throughout this paper. 
We denote a graph $\mathcal{G}$ by $(\mathcal{V}, \mathcal{E}, \mathbf{X})$ with a set of nodes $\mathcal{V} = \{v_i\}_{i = 1}^{n}$ where $|\mathcal{V}|=n$, a set of edges $\mathcal{E} \subseteq	 \mathcal{V} \times \mathcal{V}$ among these nodes that represent the connections between node pairs where $|\mathcal{E}|=m$, and the feature matrix $\mathbf{X} \in \mathbb{R}^{n \times d}$ where each row $\mathbf{x}_i \in \mathbb{R}^{d}$ is a $d$-dimension feature of node $v_i$. The topological information of graph $\mathcal{G}$ is described by the adjacency matrix $\mathbf{A} \in \mathbb{R}^{n \times n}$ where $\mathbf{A}_{ij} = 1$ if $(v_i, v_j)\in \mathcal{E}$ and $\mathbf{A}_{ij} = 0$ otherwise. Neighbors of a node $v_i$ are denoted as $\mathcal{N}(v_i)$, which consists of node $v_j$ that is connected with node $v_i$ (i.e., $\mathcal{N}(v_i) = \{v_j|(v_i,\ v_j) \in \mathcal{E}\}$). For example, in social networks, users 
are represented as nodes and their actions (e.g. commenting, following) are modeled as the edges. In recommendation, users and items are nodes and the user-item interactions (e.g. purchasing) are the edges.






\subsection{Deep Learning on Graphs}
Owing to the powerful representation ability of graphs and the rapid development of deep learning, Graph Neural Networks (GNN) have achieved impressive success in wide applications, and their powerful performances in various applications have been demonstrated~\cite{wu2020comprehensive,yue2020graph,ashoor2020graph,sawhney2021stock, hu2023adept,said2023neurograph}.
In the following part, using the basic graph notations and definitions from Section~\ref{sec:gnn-notation}, we first introduce the main idea of GNNs via the neural message-passing mechanism. Then, we provide a brief introduction to some of the most popular GNNs.

\subsubsection{Neural Message Passing GNNs}
\begin{figure}[t]
    \centering
    \includegraphics[width = .98\columnwidth]{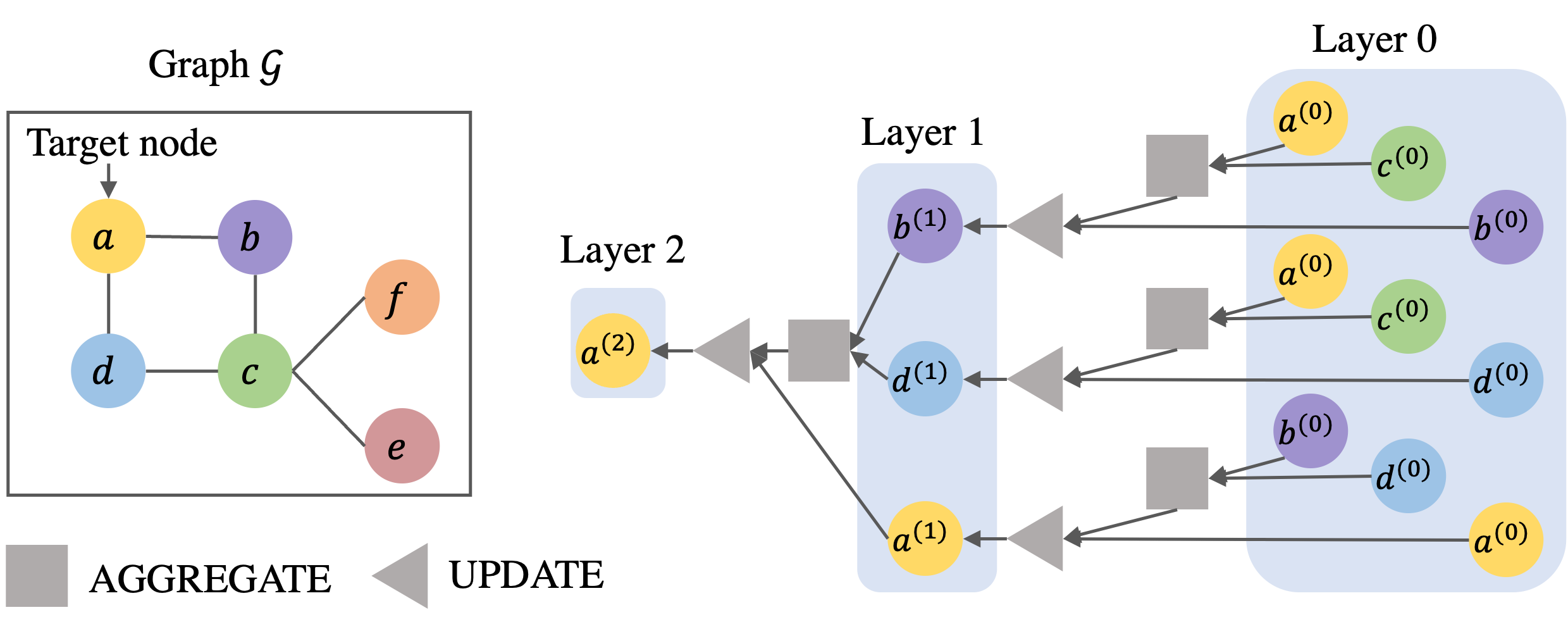}
    \caption{
    The two-step message passing framework commonly used in many GNNs: AGGREGATE and UPDATE.
    }
    \label{fig:message passing}
    \vspace{-1ex}
\end{figure}

The main idea of GNNs is to leverage the message passing mechanism, which iteratively collects information from neighbors and integrates the aggregated message with the current node representation. This iteration is described by two stage: AGGREGATE and UPDATE (illustrated in Figure \ref{fig:message passing}). In one layer, these two steps iterate for all nodes in graph $\mathcal{G}$. Stacking layers contributes to building powerful GNNs that capture higher-order relationships. We first give the message passing formula for a target node $u$ in the $k^{\text{th}}$ layer in terms of AGGREGATE and UPDATE in Eq.~(\ref{equ: aggregate}) and Eq.~(\ref{equ: update}). Then we will introduce the details of these two steps. More detailed explanations of how they are designed in different models are in Section \ref{GNN models}.
\begin{equation}
\mathbf{m}^{(k)}_{\mathcal{N}(u)} = \text{AGGREGATE}(\{\mathbf{h}^{(k)}_v, \forall{v \in \mathcal{N}(u)}\})
\label{equ: aggregate}
\end{equation}
\begin{equation}
\mathbf{h}^{(k+1)}_v = \text{UPDATE}(\mathbf{h}^{(k)}_v, \mathbf{m}^{(k)}_{\mathcal{N}(u)})   
\label{equ: update}
\end{equation}

\textbf{AGGREGATE: }Aggregation means gathering information from neighbors. At $k^{\text{th}}$-iteration
of the GNN, the AGGREGATE function takes the embeddings of the target node $u$’s neighbors $\mathcal{N}(u)$ and generates a new representation $\mathbf{m}^{(k)}_{\mathcal{N}(u)}$ based on the collected embeddings. Fundamentally, the AGGREGATE operator is a set function where the input is a set of neighbor’s embeddings and the output is a single vector. There are many choices for this aggregate operator, such as degree-based graph convolution (e.g., GCN~\cite{kipf2016semi} and  Graphsage~\cite{Graphsage}) or using attention-based strategies (e.g., GAT~\cite{GAT}) to design an AGGREGATE function.

\textbf{UPDATE: }Update refreshes the representation of a node $u$ with its own feature information and the aggregated messages from neighbors $\mathbf{m}^{(k)}_{\mathcal{N}(u)}$. At the $k^{\text{th}}$ layer of the GNN, the update function combines the new representation $\mathbf{m}^{(k)}_{\mathcal{N}(u)}$ and node $u$'s embedding $\mathbf{h}^{(k)}_u$ to obtain the new user embedding $\mathbf{h}^{(k+1)}_u$. A typical UPDATE operator involves a linear combination of the node’s embedding and the aggregated message along with a non-linear activation function (e.g., sigmoid, Tanh, ReLU) after the linear transformation. Other variants have been proposed to further improve the performance, such as concatenation methods~\cite{Graphsage} and skip-connection methods~\cite{JK}.

\subsubsection{Typical GNNs}
\label{GNN models}
Here are several widely-used GNNs along with their corresponding message-passing functions. The AGGREGATE function is denoted by $f^{(k)}_{A}$ through which we obtain the $\mathbf{m}$ representation, and UPDATE function is denoted by $f^{(k)}_{U}$. $\sigma$ denotes the activation function, and $\mathbf{W}$ and $\mathbf{a}$ are trainable parameters with their detailed dimensions illustrated later when detailing the models. 
The dimensions of node representations before/after the linear transformation are denoted as $d/d^{\prime}$, respectively.

\textbf{GCN}~\cite{kipf2016semi}.
The graph convolutional network (GCN) is one of the most popular GNNs with aggregation and update functions as follows:
\begin{equation*}
    f^{(k)}_{A}(\{\mathbf{h}_{v}^{(k-1)} | v \in \mathcal{N}(u)\}) = \sum_{v \in \mathcal{N}(u)}{\frac{\mathbf{h}_{v}^{(k-1)}}{\sqrt{deg(u)deg(v)}}}
\end{equation*}

\vspace{-0.5ex}
\begin{equation*}
    f^{(k)}_{U}(\{\mathbf{h}_{u}^{(k-1)}, \mathbf{m}^{(k)}_u\}) = \sigma(\mathbf{W}^{(k)}\mathbf{m}^{(k)}_u)
\end{equation*}

\noindent where the aggregation is normalized by degree (i.e., $deg(\cdot)$) from both source and target nodes and $\mathbf{W}^k \in \mathbb{R}^{d^\prime \times d}$ is the weight of the linear transformation layer.

\textbf{GraphSAGE}~\cite{Graphsage}.
GCN models are inherently transductive. They cannot efficiently generalize to unseen nodes or different graphs. GraphSAGE is proposed to solve this issue in an inductive manner, which samples a fixed-size neighborhood of each node and leverages node attribute information to efficiently generate representations on previously unseen data. $\oplus$ operation concatenates the embedding of the target node and aggregated message, and $\mathbf{W}^k$ has dimension of $\mathbb{R}^{d^\prime \times 2d}$ due to the concatenate operation.
\begin{equation*}
    f^{(k)}_{A}(\{\mathbf{h}_{v}^{(k-1)} | v \in \mathcal{N}(u)\}) = \frac{1}{deg(u)}\sum_{v \in \mathcal{N}(u)}{\mathbf{h}_{v}^{(k-1)}}
\end{equation*}

\vspace{-1ex}
\begin{equation*}
    f^{(k)}_{U}(\{\mathbf{h}_{u}^{(k-1)}, \mathbf{m}^{(k)}_u\}) = \sigma(\mathbf{W}^{(k)}[\mathbf{h}_{u}^{(k-1)} \oplus \mathbf{m}^{(k)}_u])
\end{equation*}

\textbf{GAT}~\cite{GAT}.
Graph Attention Networks (GAT) introduce attention mechanism to compute the weight between edge $e_{uv}$ (denoted as $\alpha_{uv}$), which indicates that different neighbors will contribute differently in the aggregation process based on the learned node representations where $\mathbf{W}^k \in \mathbb{R}^{d^\prime \times d}$ and $\mathbf{a} \in \mathbb{R}^{2d^\prime \times 1}$.
\begin{equation*}
    \alpha^{(k)}_{uv}=\frac{\exp{(\sigma(\mathbf{a}^{(T)}[\mathbf{W}^{(k)}\mathbf{h}_{u}^{(k-1)} \oplus \mathbf{W}^{(k)}\mathbf{h}_{v}^{(k-1)}]))})}{\sum_{v^{'} \in \mathcal{N}{(u)}}\exp{(\sigma(\mathbf{a}^{(T)}[\mathbf{W}^{(k)}\mathbf{h}_{u}^{(k-1)} \oplus \mathbf{W}^{(k)}\mathbf{h}_{v^{'}}^{(k-1)}]))}}
\end{equation*}

\vspace{-0.5ex}
\begin{equation*}
    f^{(k)}_{A}(\{\mathbf{h}_{v}^{(k-1)} | v \in \mathcal{N}(u)\}) = \sum_{v \in \mathcal{N}(u)}{ \alpha^{(k)}_{uv} \mathbf{h}_{v}^{(k-1)}}
\end{equation*}

\vspace{-0.5ex}
\begin{equation*}
    f^{(k)}_{U}(\{\mathbf{h}_{u}^{(k-1)}, \mathbf{m}^{(k)}_u\}) = \sigma(\mathbf{W}^{(k)}\mathbf{m}^{(k)}_u)
\end{equation*}

We encourage readers who seek a more comprehensive introduction to deep learning on graphs to explore dedicated surveys~\cite{wu2020comprehensive,zhou2020graph}, tutorials~\cite{rong2020deep,wu2022graph} and books~\cite{wu2022graph,ma2021deep}.

\subsection{Motivation}
Although data privacy has been well-investigated in the general DL domain, it is rather critical to consider privacy for graph data and models since (1) graph data and models are prevalent in real-world applications; (2) the data and models are vulnerable to attacks, it is less explored than other forms of data (e.g., tabular);  
(3) extending directly from regular data to graph data without any domain modification is challenging due to the complex graph-based structure. Therefore, discussing privacy attacks and preservation techniques that are specific to graphs becomes a necessity~\cite{sun2022adversarial}.

\subsubsection{Vulnerabilities of Graph Data and GNN Models}
Compared to regular tabular, image, and text data, graph data has complex connections and potential edge features, both of which could be considered as potential objects to attack~\cite{yuan2023navigating, zhang2022chasing}. 
For example, while most other forms of data are independent and identically distributed (i.i.d.), the nodes within a graph are inherently connected. Furthermore, many real-world networks have high homophily~\cite{mcpherson2001birds} where connected nodes are prone to share similar features. Thus, this can be exploited to infer the node information based on its neighborhood and cause the risk of information leakage, which poses brand new challenges to privacy preservation. However, we note that even for graphs not exhibiting high homophily, GNNs have shown to still perform well with lower levels of homophily~\cite{ma2021homophily,luan2023graph}, which suggests they are also vulnerable to privacy attacks. 

Furthermore, the complex connections among nodes make it hard for data partitioning and thus bring huge challenges for distributed training. Therefore, GNNs are typically trained in a centralized way where the model and data are stored in one place, which increases the risk of information leakage and might be impossible in many real-world settings. While GNNs have shown significant improvement in various applications, the unique message-passing process might exacerbate the sensitive leakage as each node now encodes the information from its neighbors. It means that to protect one node, not only that specific node should be protected as in i.i.d data, but the substructures surrounding that node should also be protected.

\subsubsection{Related Surveys and Differences}

Recently, several surveys~\cite{dai2022comprehensive,wu2022survey, zhang2022trustworthy, khosla2022privacy} have been conducted on the trustworthiness in GNNs, including the reliability, explainability, fairness, privacy, and transparency aspects, showing the common interest and concern on trustworthiness with our interests here.
On one hand, within the range of trustworthiness, although surveys related to explainability~\cite{yuan2022explainability}, and fairness~\cite{chen2023fairness} of GNNs exist, few of them provide a comprehensive and focused discussion on privacy for GNNs. On the other hand, although there are privacy surveys in ML/DL~\cite{boulemtafes2020review, mireshghallah2020privacy, rigaki2020survey} and social networks~\cite{beigi2020survey,kayes2017privacy}, privacy specific to graph data and graph models has not been comprehensively discussed yet. This motivates us to present a systematic and in-depth review of the existing attack models and privacy-preserving techniques on GNNs, which could benefit the research community in developing privacy-preserving GNNs immune to privacy attacks. 

\section{Privacy attacks on GNNs} \label{sec:priv_gnn}
\label{sec:attack}

Privacy attack is a popular and well-developed topic in various fields such as social network analysis, healthcare, finance, system, etc. ~\cite{yu2022pas,  hussain2021security, chong2021privacy}. During recent years, the surge of machine learning has provided powerful tools to solve many practical problems. However, data-driven approaches also threaten users' privacy due to the associated risks of data leakage and inference~\cite{mireshghallah2020privacy}. Consequently, a substantial amount of work has been devoted to investigate the vulnerabilities of ML models and the risks of privacy leakage~\cite{rigaki2020survey}. A branch of privacy research is to develop privacy attack models, which has received much attention during the past few years. However, attack models with respect to GNNs have only been explored very recently because GNN techniques are relatively new compared with CNN/transformers in image/natural language processing(NLP) domains,
and the irregular graph structure poses unique challenges to transfer existing attack techniques that are well-established in other domains. In this section, we summarize papers that have developed attack models specifically targeting GNNs.

\begin{figure}[t]
    \centering
    \includegraphics[width = .98\columnwidth]{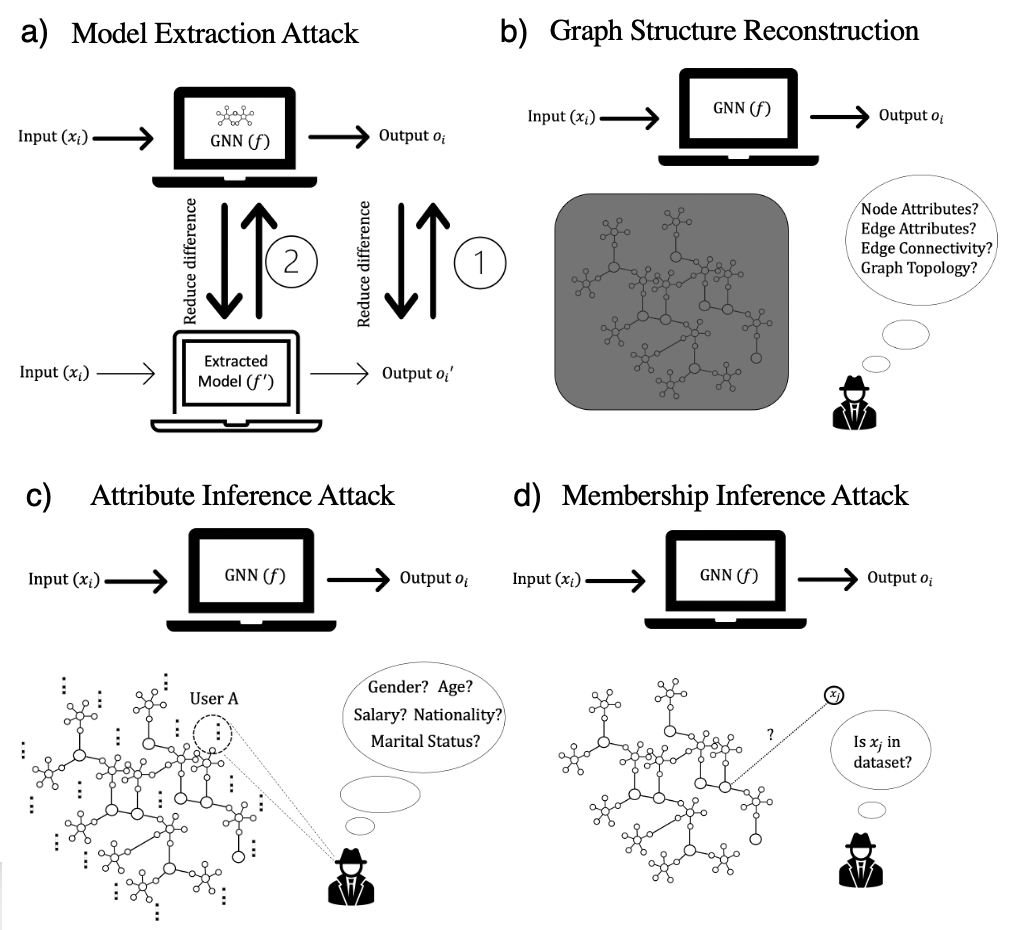}
    \vspace{1ex}
    \caption{Illustrations of the four categories of privacy attack models on graphs: a) Model extraction attacks (MEA); b) Graph structure reconstruction (GSR); c) Attribute inference attacks (AIA); and d) Membership inference attacks (MIA). }
    \label{fig:mia}
\end{figure}

We classify the privacy attack models on GNN into four categories (which are visualized in Figure~\ref{fig:mia}): a) model extraction attack (MEA), b) graph structure reconstruction (GSR), c) attribute inference attack (AIA), and d) membership inference attack (MIA). 

In MEA, the GNNs model is often directly extracted/inferred with the aid of a surrogate model. Concretely, the surrogate model is trained so that it can output predictions similar to the ground-truth values that would be generated by the target model given the same input. In GSR attack, information related to graph structure such as topology and connectivity is inferred by the attackers. Compared to MEA, GSR aims to obtain more information about the model rather than simply mimicking the performance. GSR is similar to the model inversion attack mentioned in the previous section except it is for graph specifically. 
Note that GSR is equivalent to graph information reconstruction (GIR) that is used in the literature. 
We rename it to GSR because this attack focuses more on the reconstruction of graph structural information. In AIA, concrete features of node (e.g. age, salary) are obtained by the attackers. In MIA, the attackers aim to determine whether or not a node belongs to the training set. Another way to categorize the attack models is based on the accessibility of information, and the models can be divided into \textit{white-box attacks} and \textit{black-box attacks} respectively. In the white-box setting, adversaries are assumed to be able to access rich information of GNNs such as their architecture, parameters, embeddings, and outputs. By contrast, in black-box settings, adversaries have limited information about the target model, if not none.

\subsection{Model Extraction Attack}

MEA often occurs in models based on multi-layer perceptron (MLP) and convolutional neural networks (CNNs). Recently, a trend of increasing popularity in researching MEA on GNN has been observed. Under the MEA scheme, the attackers typically have limited information about the GNN model (i.e., black-box). During a MEA, the attackers first adopt a model (e.g., GNN) similar to the victim GNN model. Subsequently, the adopted model is tuned so that it has a similar performance as the target model in terms of criteria such as accuracy and decision boundary. To accomplish this, the attackers first generate queries to the victim model, then collect the outputs from the model API, and subsequently feed the same queries to the extracted model, and finally tune the parameters of the adopted model so that it can have similar outputs as does the victim model. As the performance of the two models converges, the adopted model can be considered as an extracted model of the victim model. The mathematical definition of MEA to GNN-based approach is the following: a GNN model can be expressed as $f$ operated on a graph $G$. The goal of MEA is to obtain an extracted model $f^{\prime}$ such that $f^{\prime}(G)\approx f(G)$. Note that due to the connectivity of the data samples in a graph, MEA to GNN will be facilitated if additional information about the graph structure is available. This is one distinct difference from the MEAs to CNN-based or MLP-based models, in which the data samples are not connected, unlike the connected nodes in a graph.

An early work studying MEA towards GNN is done by \cite{defazio2019adversarial}, where the extracted model is able to produce the output with an $80\%$ similarity to that of the victim model (also called fidelity) on Cora and Pubmed datasets, with the limited access to only a subgraph.
A recent comprehensive investigation of MEA towards GNNs is done by \cite{wu2022model}. The authors present seven possible scenarios under which the attacker possesses a different amount of prior information (e.g., node attributes and topology) for the attack. Based on these scenarios, they determine seven categories of attack. After experimenting with their attack on multiple real-world datasets, they claim that their extracted model can reach a fidelity of as high as $90\%$.
Similarly, other recent work has also focused on the MEA on inductive GNNs~\cite{shen2022model}. 



\subsection{Graph Structure Reconstruction}
In this subsection, we will discuss recent progress about GSR specifically targeting GNN. In a GSR attack on GNN, the attackers seek to steal the private information of the input graph, mainly pertinent to the graph structure.


Duddu et al.~\cite{duddu2020quantifying} conduct a GSR on GNN model to extract the graph $G_{\text {target}}$ using the publicly accessible node embedding $\Psi(v), \forall v \in G_{\text {target }}$. The process has 2 phases, of which the first one trains a graph-encoder-decoder structure using the auxiliary graph $G_{\text {auxiliary}}$ of which nodes follow the data distribution of the target graph $G_{\text {target}}$, and the second phase leverages the publicly accessible target node embedding and the trained decoder to estimate the target adjacency matrix $A_{\text{target}}$, which captures the graph connectivity and edge distribution. They demonstrate that their proposed GSR reaches a very high precision (above 0.7) for estimating the target graph using Cora, Citeer, and Pubmed.
Zhang et al.~\cite{zhang2022model} look at GSR for edge reconstruction and structural reconstruction, where they call their attack model Graph Module Inversion Attack (GraphMI). In the white-box attack setting, GraphMI consists of a projected gradient module, a graph auto-encoder module, and a random sampling module. The projected gradient module aims to extract graph topology with the output labels of the target model and auxiliary knowledge, and is designed to tackle discrete optimization problems via convex relaxation while preserving graph sparsity and feature smoothness. Then, the graph auto-encoder module takes node attributes, graph topology, and target model parameters into consideration for graph reconstruction. Specifically, the projected gradient module solves the following optimization problem:
\begin{equation*}
    \label{equ: LinkInfer1}
    \mathop{\arg\min}_{\mathbf{a}\in \left[0,1\right]^n }\mathcal{L}_{GNN}+\alpha\mathcal{L}_{s}+\beta\|\mathbf{a}\|_2
\end{equation*}
where $\mathbf{a}\in\left[0,1\right]^n (n=N(N-1)/2)$ is a continuous-valued adjacency vector, which is transformed from the adjacency matrix $\mathbf{A}\in\{0,1\}^{N\times N}$ so that the original combinatorial optimization problem can be solved through the projected gradient descent method, $\mathcal{L}_{GNN}$ denotes the loss of the target model aiming to make the reconstructed adjacency more similar to the original one, $\mathcal{L}_s$ is the term to ensure the feature smoothness in the optimized graph, the last term is to encourage the sparsity of graph structure, and $\alpha, \beta$ are constant parameters. One can refer to \cite{zhang2021graphmi} for details of the model. The graph auto-encoder module is composed of an encoder and a decoder. The encoder is directly transferred from the target model $f(\mathbf{x};\mathbf{\theta}^*)$ with partial parameters (i.e., excluding the readout layer). After the optimization of projected gradient module, the encoder encodes nodes to node embeddings by using topology information and node attributes $\mathbf{x}$. Then, the decoder will reconstruct the graph adjacency based on the node embeddings. The random sampling module is designed to recover the binary adjacency matrix. 

In \cite{zhang2021graphmi}, the authors claim that GraphMI is effective towards inferring edges after evaluating it on three GNNs (i.e., GCN~\cite{kipf2016semi}, GAT~\cite{GAT}, GraphSAGE~\cite{Graphsage}). Further, based on their analysis of the relation between the edge influence and the model inversion risk, they conjectured that the ease of reconstruction is positively correlated with the edges' influence. In addition, the authors incorporate gradient estimation and reinforcement learning into GraphMI to render it capable for black-box attack. 

He et al. \cite{he2021stealing} first propose a threat model called link stealing attack aiming to infer the existence of links among nodes in target datasets. The authors systematically characterize the background knowledge (i.e., accessibility of knowledge) of an adversary through three dimensions, which are attributes of nodes in the target datasets $\mathbf{X}$, partial graph structure in target datasets $\Bar{\mathbf{A}}$, and a shadow dataset ${D}_{Shadow}$, respectively. Based on different accessibility settings (i.e., have access or not) of the knowledge in three dimensions, they develop in total 8 attack mechanisms for all 8 possible settings. Basically, if an adversary has no knowledge or only the target node attributes $\mathbf{X}$, the attack model is conducted in an unsupervised way, which is mainly based on the intuition that a node pair would share more similar attributes or closer posteriors queried from the target model when linked. If an adversary has access to the partial graph structure in target datasets $\Bar{\mathbf{A}}$ or a shadow dataset ${D}_{Shadow}$, the attack model can be trained in a supervised way. For partial graph $\Bar{\mathbf{A}}$, the adversary takes the links as the ground truth label to train an attack model. Also, an adversary could train an attack model based on a shadow dataset ${D}_{Shadow}$ so that the trained model can be later transferred to infer links in the target dataset. The intuition is that the trained model could obtain the ability to capture the similarity between two nodes’ posteriors, which can be further transferred to different datasets (e.g., target model). One can refer to \cite{he2021stealing} for more details about each attack model. Plenty of experimental results demonstrate the effectiveness of the proposed attack models. More importantly, the results indicate that the output predictions of GNNs preserve rich information about the structure of a graph that is used to train the model. Wu et al.~\cite{wu2022linkteller} also focus on edge privacy and aim to recover private edges from GNNs through influence analysis. \cite{olatunji2022private} demonstrates that additional knowledge of post-hoc feature explanations substantially enhances the structural attacks.

\subsection{Attribute Inference Attack}
AIA aims to infer the properties of a target training dataset. For graph-based data, the properties are usually related to nodes and edges, and these properties could be sensitive features (e.g. gender and age information of a user node in social network analysis~\cite{olatunji2023does}, chemical bonds information in a molecular graph). Note that the difference between GSR and AIA can be subtle. The former one aims to reproduce the graph, while the later one focuses on stealing the concrete features of the dataset. Compared to many other adversarial attacks, AIA is often considered to be more malicious due to its capability to directly predict the sensitive features of a target user. In addition, an even more serious derivative of AIA, data reconstruction attack, could occur when attackers try to infer a subset of the training data, rather than a single one. In AIA, embedding is often used to predict the sensitive features because of their close relationship. 

Duddu et al.~\cite{duddu2020quantifying} investigate AIA on graphs by inferring the gender and location of certain users of the targeted graphs composed with Facebook and LastFM. In their study, a fraction of users publicly disclose their gender and location, and a sub-graph $G_{\text{aux}}$ can be constructed based on these users. Here, given $\left(\Psi(v), s_v\right) \forall v \in G_{\text{aux}}$, where $\Psi(v)$ is node embedding and $s_v$ is the disclosed sensitive feature, a supervised attack classifier model $f_{\text {attack}}$ can be trained. Based on the trained model and the available node embeddings of the target graphs, their sensitive features can be estimated via $f_{\text {attack}}\left(\Psi\left(v^{\prime}\right)\right)\ \text{where}\ v^{\prime} \in G_{\text {target}}$. 
Their AIA showed high performance. 

Although less common, edges could also be related to or contain sensitive information. For example, in a drug-target interaction graph, a link could contain sensitive information such as the interaction pattern and affinity between certain drug and target. Thus, AIA could also be designed specifically against the edges.

\subsection{Membership Inference Attack} 
Membership is to describe whether a data sample belongs to the training dataset or not~\cite{li2021membership}. In GNNs, the goal of a MIA can be at the node level, edge level, and graph level~\cite{duddu2020quantifying,olatunji2021membership,he2021node,conti2022label,he2021stealing,wu2021adapting}. Node-level MIA is to infer whether a node exists in the original graph or not. For instance, attackers could be interested in whether or not a person is in a certain social community by using an MIA on that social network; subsequently, the attackers could further modify/steal the private information of the targeted user. Edge-level MIA aims to infer the membership of links. Graph-level MIA is to infer whether a graph is used during the training which is often used for graph classification tasks. Typical methods for MIA include using shallow models, node embedding, and graph embedding~\cite{xu2023modeling,zhao4441020mia}. MIA is a binary classification problem, and typical metrics for evaluating MIA include the inference accuracy, precision (i.e., the percentage of true positive), ROC-AUC score enabling the visualization of true positive rate versus false positive rate, etc.\\

\textbf{Node-level MIA.} 
In \cite{olatunji2021membership}, the authors perform MIA with the help of a shadow model (i.e., a simplified model trained to approximate the target model), where they build the attack model in a supervised way with data generated from the trained shadow model, under the assumption that they can construct the shadow model using the same neural architecture as the target model. They train the shadow model with data from the same distribution as the training set of the target model. In order to better simulate the behaviors of the target model, the output probabilities queried from the target model are considered as the ground truth during the training of the shadow model. The attack model is designed as a binary classification model, which maps the output prediction confidence of a model to the membership of the corresponding input nodes. To train the attack model through a supervised way, the authors generate the dataset by using the trained shadow model to predict on the entire dataset $\mathcal{D}_{Shadow}$ (both member and non-member nodes) and obtain the corresponding output confidence. Finally, they can infer the membership with the trained attack model once they have the output of a node queried from target models. In addition, they also try another way to obtain the shadow model. Instead of querying the target model for confidence, they use ground truth labels of the original nodes to train the shadow model. Interestingly, it is found that there is no significant difference in the attack success rate. The authors also claim that it is unnecessary for the shadow model to have exactly the same architecture as does the target model after they realize that a standard graph convolutional network would also gain good results in MIA.

In \cite{duddu2020quantifying}, the authors develop the MIA method under black-box and white-box settings respectively. In the black-box setting, an adversary is assumed to have only access to the output probabilities of the target model when given a node. Thus, under this scenario, the authors consider exploiting the statistical difference between the prediction confidence on training (i.e.) member and non-member data). Specifically, they demonstrate that if a node belongs to the training data, the output probability queried from the target model would be more confident (i.e., has higher values) on the corresponding label. While for a non-member node, the output probability distribution is supposed to be less confident and more uniform. Then, based on this setting, the authors consider two attack methods (i.e. shadow attack and confidence attack) respectively. The shadow model shares the similar idea as does the one in \cite{olatunji2021membership}, which builds the attack model through a supervised way with the training data generated from the trained shadow model. By contrast, a confidence attack performs inference in an unsupervised setting. Based on the fact that nodes with a higher prediction confidence are more likely to be members, an adversary could decide memberships according to whether the highest confidence of a node’s prediction is above a certain threshold which can be set or learned. The authors also experimentally verify that the confidence attack performs much better than does shadow attack under the black-box setting. Then, in the white-box setting, an adversary has the access to the intermediate output of the target model (i.e., node embedding in \cite{duddu2020quantifying}). Here, the authors propose an unsupervised method that maps the intermediate embedding to a single membership value. Specifically, they train an encoder-decoder model. The encoder encodes the intermediate embeddings with a single membership value, which is then passed to the decoder to reconstruct the embeddings. Afterwards, they use clustering method (e.g., K-Means) to distinguish the obtained single membership value into the clusters (i.e., members and non-members). The authors conducted their analysis using a wide range of GNNs and network-embedding methods (e.g. GCN, GraphSAGE, GAT, Topological Adaptive GCN, DeepWalk, Node2Vec) and thus show the generalizability of their result.

He et al. \cite{he2021node} propose a MIA model also based on training a shadow model. Instead of first querying from the target model for the output probabilities of nodes in $\mathcal{D}_{Shadow}$ as \cite{olatunji2021membership}, the authors directly train the shadow model with the dataset $\mathcal{D}_{Shadow}$ as well as the corresponding features and labels, which are derived from the same distribution as the training set $\mathcal{D}_{Target}$ of the target. Similar to \cite{olatunji2021membership}, the attack model is trained through a supervised way with membership information in $\mathcal{D}_{Shadow}$ and the queried posteriors of nodes from the trained shadow model. Differently, depending on the adversary's knowledge of node topology, the authors develop three query methods, which are 0-hop query, 2-hop query, and combined query, respectively. Also, there are three different attack models that are trained with datasets corresponding to three query methods respectively. The experimental results demonstrate that the combined attack outperforms 0-hop attack and 2-hop attack, since it takes advantages of the later two methods. Moreover, they experimentally show that the assumption of the identical distributions of $\mathcal{D}_{Shadow}$ and $\mathcal{D}_{Target}$, and the same architecture of the shadow model and target model can both be relaxed, which is consistent with the results in \cite{olatunji2021membership}. \\

\vspace{-0.5ex}
\textbf{Edge-level MIA.} 
Similar to the node, which contains sensitive information, the connections also carry valuable information and thus become a target of the attackers. Edge-level MIA aims to determine whether there is a link between two nodes in the training graph~\cite{he2021stealing}.\\

\vspace{-0.5ex}
\textbf{Graph-level MIA.} 
In addition to the node and edge level, researchers have also investigated graph-level membership inference~\cite{wu2021adapting, zhang2022inference,wang2021membership} where the task is to infer whether a graph/sub-graph is used in the training set. Wu et al.~\cite{wu2021adapting} aim to infer whether a graph sample has been used for training and design two types of attacks including training-based attacks and threshold-based attacks from different adversarial capabilities. Zhang et al.~\cite{zhang2022inference} infer the basic graph properties and the membership of a sub-graph based on graph embedding.

\section{\hspace{-1ex}Privacy preservation techniques on GNNs}
\label{sec:preservation}

After discussing possible attacks toward GNNs model, we now shift our attention to preservation methods that can effectively protect against these attacks. In the following subsections, we will discuss latent factor disentangling, adversarial training, differential privacy approach, and federated learning. Latent factor disentangling aims to remove the sensitive information from the embedding while minimizing the loss of meaningful information for the downstream tasks; adversarial training aims to render a model resistant to privacy attacks through careful training; differential privacy approaches incorporate random noise into data samples or intermediate model variables to protect sensitive information during queries; and federated learning enables collaboration among users with private datasets without revealing them. Table \ref{tablemethods} enumerates the techniques associated with these four categories.

\begin{table}[tbp]
    \footnotesize
    \centering
    \caption{Privacy preservation techniques that have either been utilized or have the potential to be employed on GNNs. Public code links to these methods are also provided (if available). All the methods are collected \href{https://github.com/NDS-VU/awesome-gnn-privacy}{here}.}
    \label{tablemethods}
    \begin{tabular}{ c|c|c}
\hline
Category  & Method & Public Code \\ \hline

\multirow{4}{*}{Latent Factor Disentangling}   & APGE~\cite{li2020adversarial} & \href{https://github.com/KaiyangLi1992/Privacy-Preserving-Social-Network-Embedding}{Link}\\ 
& Wang et al.~\cite{wanglearning} & -\\& DP-GCN~\cite{hu2022learning}& \href{https://github.com/HuiHu1/Privacy-Preserving-Graph-Convolutional-Network}{Link}\\&DGCF~\cite{wang2020disentangled} &\href{https://github.com/xiangwang1223/disentangled_graph_collaborative_filtering}{Link}\\ \hline 

\multirow{7}{*}{Adversarial Training}   & NetFense~\cite{hsieh2021netfense} & \href{https://github.com/ICHproject/NetFense/}{Link} \\ & Tian et al.~\cite{tian2021k} & - \\ &SecGNN~\cite{wang2023secgnn} & \href{https://github.com/songleiW/SecGNN}{Link}\\ &FRFC~\cite{liu2022dual} &-\\ &GAL~\cite{liao2021information}&\href{https://github.com/liaopeiyuan/GAL}{Link} \\ & Wang et al. ~\cite{wang2021privacy}&-\\&AttrOBF~\cite{kumar2020adversary}&-\\\hline 

\multirow{8}{*}{Differential Privacy}  
& LPGNN~\cite{sajadmanesh2020locally} & \href{https://github.com/sisaman/LPGNN}{Link} \\ 
& GAP~\cite{sajadmanesh2022gap} & \href{https://github.com/sisaman/GAP}{Link} \\ 
& Mueller et al.~\cite{mueller2022differentially} & - \\ 
& DP-Adam~\cite{daigavane2021node} & \href{https://github.com/google-research/google-research/tree/master/differentially_ private_gnns}{Link} \\ & PrivGNN~\cite{olatunji2021releasing}&-\\
& DP-GNN~\cite{mueller2022differentially1} &\href{https://github.com/tamaramueller/DP-GNNs}{Link}\\ &Solitude~\cite{lin2022towards}&-\\&GERAI~\cite{zhang2021graph}&-\\\hline

\multirow{12}{*}{Federated Learning}    
&Fedgraphnn~\cite{he2021fedgraphnn}&\href{https://github.com/FedML-AI/FedML/tree/master/python/app/fedgraphnn}{Link}\\ 
& FedGL ~\cite{chen2021fedgl}&-\\
& FedGNN ~\cite{wufedgnn2021}&-\\
& FedGCN ~\cite{yao2022fedgcn}&\href{https://github.com/yh-yao/FedGCN}{Link}\\
& VFGNN ~\cite{junver2020}&-\\
& SGNN ~\cite{mei2019sgnn}&-\\
& FedVGCN ~\cite{ni2021vertical}&-\\
& D-FedGNN ~\cite{pei2021decentralized}&- \\ & SpreadGNN ~\cite{he2021spreadgnn}&\href{https://github.com/FedML-AI/SpreadGNN}{Link}\\ & FedPerGNN~\cite{huang2022federated} &\href{https://github.com/wuch15/FedPerGNN}{Link}\\ 
& FLIT(+) ~\cite{zhu2022federated} & \href{https://github.com/ur-whitelab/fedchem}{Link}\\
& GraFeHty~\cite{sarkar2021grafehty} & - \\ \hline
    \end{tabular}
    \vspace{-2ex}
\end{table}

\vspace{-0.5ex}
\subsection{Latent Factor Disentangling}
Graph/node embedding is able to preserve sensitive and non-sensitive information of the original graph. From the perspective of representational learning, latent factor disentangling aims to disentangle the sensitive information from the embedding while ensuring the utility of the disentangled embedding for downstream tasks. In this way, it would be difficult for an adversary to implement privacy attacks with limited/no latent sensitive information. Various methods have been proposed to achieve this goal~\cite{li2020adversarial, wanglearning, hu2022learning}. 

Li et al.~\cite{li2020adversarial} propose a graph embedding model to disentangle sensitive information from the embedding while preserving the structural information and data utility. The model incorporates two mechanisms in a complementary way, both of which can separately protect the sensitive information. The first mechanism is based on the graph autoencoder (GAE) model \cite{kipf2016variational}, which is able to process graph-structured data, or a supervised Adversarial Autoencoder (AAE) model \cite{makhzani2015adversarial}, which augments the decoder with the one-hot encoding of the label so that the final embedding would contain label-invariant information. According to this, the authors use GCN as encoder, and incorporate privacy labels to the decoder to disentangle sensitive information from the final embedding. The proposed autoencoder achieves graph reconstruction with the output containing both link prediction and the prediction of non-sensitive node attributes. Note there also has a discriminator for updating the encoder. The loss function can be described as $\mathcal{L}_{recon}=\mathcal{L}_{link}+\mathcal{L}_{attr}$
where $\mathcal{L}_{link}$ and $\mathcal{L}_{attr}$ denote loss for link prediction and the prediction of non-sensitive node attributes respectively. The discriminator has a separate loss function $\mathcal{L}_{dc}$. 

Instead of disentangling privacy labels as decoder input, 
the second mechanism achieves the similar effect by incorporating an attacker model (i.e., a softmax classifier) that aims to predict sensitive node attributes to an obfuscator (i.e. similar to the autoencoder model of the first mechanism while without incorporating privacy labels). The goal of the second mechanism is to reduce the performance of the attacker while preserving the graph structure and non-sensitive data utility. Given the loss function of the attacker $\mathcal{L}_{attack}$, the loss function of the obfuscator will be $\mathcal{L}_{obf}=\mathcal{L}_{recon}-\lambda\mathcal{L}_{attack}$
where $\lambda$ is a trade-off hyper-parameter. Finally, the authors use 
two mechanisms together to 
protect the privacy in both input and output of the decoder in a complementary way.

Wang et al.~\cite{wanglearning} design a framework based on encoder-decoder architecture to learn graph representations that encode sufficient information for downstream tasks while disentangling the learned representation from privacy features. The key component is a conditional variational graph autoencoder (CVGAE), which captures the relationship between the learned embeddings and the sensitive features. They add a penalty loss into the original reconstruction objective to encourage the CVGAE to minimize the mean Gaussian distribution differences so that the privacy leakage will be punished. Under this framework, they consider two specific scenarios where the graph structure is available and unavailable to the adversary, respectively.

In some graph datasets, the same attribute could be private for some nodes, while non-private for the other nodes. One example is social network composed with people of different genders, where the age of people of certain genders (e.g. female) could be private while the age of the people of the rest of genders (e.g. male) could be public. Due to graph homophily (i.e. connected nodes are similar), showing the public information about some users (e.g. ages of men) could lead to inference about some private information (e.g. ages of women who are connected to the men on the social network). To handle this issue, Hu et al.~\cite{hu2022learning} propose a privacy-preserving GCN model named DP-GCN that can conceal the value of the private sensitive information that has been exposed by public users in the same network. DP-GCN has a Disentangled Representation Learning Module (DRL) and a Node Classification Module (NCL). DRL separates the non-sensitive attributes into sensitive and non-sensitive latent representations 
to be orthogonal to each other. NCL trains the GCN to determine the class of the nodes with non-sensitive latent representations. The authors experimentally showed that the attributes disentangling of the public users (e.g. men with disclosed ages) can help protect the privacy of the private users (e.g. women with undisclosed ages). 

Besides the above influential works, researchers have also worked in this direction~\cite{wang2020disentangled}, introducing innovations that improve performance and inspire further research in privacy-preserving.



\subsection{Adversarial Training}
An intuitive perspective to defend against privacy attacks is to directly reduce the performance of specified attacks. Accordingly, one can train models with the objective of minimizing the performance of specified privacy attacks (i.e., privacy protection) and maintaining the performance of downstream tasks (i.e., utility). We refer to this kind of approach as adversarial training. Actually, the aforementioned model in \cite{li2020adversarial} can also be categorized to adversarial training. Specifically, the updating process contains loss functions of the discriminator $\mathcal{L}_{dc}$ and the attacker $\mathcal{L}_{attack}$, which will make the model robust to the attack in an adversarial way.

Hsieh et al. \cite{hsieh2021netfense} propose an adversarial defense model against GNN-based privacy attacks named NetFense, which is able to reduce the prediction performance of the attacker and maintain the data and model utility (i.e., maintain the performance on downstream tasks). Different from \cite{li2020adversarial}, this paper presents a graph perturbation-based approach aiming to make perturbation to the original graph (i.e., changing the adjacency matrix, $\mathbf{A}$) to fool the attacker while maintaining utility for the downstream task (i.e., node classification). To achieve this, the model comprises three phases, which are candidate selection, influence with GNNs, and combinatorial optimization, respectively. The first phase aims to find out a set of candidate edges for perturbation. Specifically, it uses the change of Personalized PageRank (PPR) score to measure the noticeability of the perturbation. To keep data utility, it selects candidate edge perturbation with minimal noticeability. Then, the second and third phases ensure model performance and privacy preservation. Compared to \cite{li2020adversarial}, which alternately updates the model and attacker, NetFense does not update models in an adversarial way. Instead, given the pre-trained classification model and attack model, NetFense modifies the input graph structure to obtain a perturbed graph that can protect privacy. Another difference is that Netfense assumes privacy labels are binary. Thus, it aims to reduce the prediction accuracy of the attacker to 0.5 rather than just maximizing the loss function of the attacker.

 
Tian et al.~\cite{tian2021k} adopt adversarial training for privacy preservation for social network analysis. Concretely, their model has two stages, of which the former one is based on an $\in$-k anonymization method, and the later one is based on an adversarial training mechanism. The adversarial training can help GNN extract useful information from the anonymous social network data after the first step. In other words, the main purpose of adding the adversarial training is to render GNNs resistant to the disturbance from $\in$-k anonymization so that the privacy can be enhanced and the model performance is minimally compromised.
The authors also showed the effectiveness of their approach through experiments related to classification, link prediction, and graph clustering using multiple real-world datasets. Further, the authors showed the flexibility and efficiency of their model in the data collection/training phases.  

Apart from the aforementioned and discussed influential studies in this area, additional researchers have also explored this field~\cite{liu2022dual,liao2021information,wang2021privacy,kumar2020adversary}, introducing various innovations that have led to increased performance and inspiring even more future work.


\vspace{-0.5ex}
\subsection{Differential Privacy} 
As discussed in section \ref{sec.privacy_and_attacks_DL}, differential privacy has been successfully applied to various machine learning models for privacy protection. However, unlike other machine learning models wherein the data points are usually assumed to be independent, the data samples of GNNs are nodes of graphs, which are connected by edges, making the methods developed for other machine learning models difficult to be directly applied to GNNs. Recently, increasing works develop differentially private (DP) approaches to solve privacy preserving problem particularly for GNNs. The key point of DP approaches is to add random noise to data samples or intermediate model variables so that when querying for the sensitive information, an adversary would only obtain perturbed data, which is useless for attacks. Obviously, data with too much noise will also degrade the performance of GNNs, making it a major issue to find a good balance between data privacy and utility.

Sajadmanesh et al.~\cite{sajadmanesh2020locally} present a locally private GNN (LPGNN) using the local differential privacy (LDP) technique to protect data privacy. They add noise to both node features and labels to ensure privacy and design denoising mechanisms so that LPGNN is able to be trained with the perturbed private data. More specifically, to add noise on node features, they propose an LDP encoder and an unbiased rectifier, where the former is to inject noise (i.e., $\mathbf{X}\rightarrow\mathbf{X}^*$), and the latter converts the encoded vector $\mathbf{X}^*$ to an unbiased perturbed vector $\mathbf{X}^{'}$ (i.e., $\mathbb{E}[\mathbf{X}^{'}]=\mathbf{X}$). Then, to denoise the perturbed node features, they append a multi-hop aggregation layer to the GNN, called KProp, which is able to average out the injected noise. To perturb node labels, the authors use the generalized randomized response mechanism \cite{kairouz2016discrete}. LPGNN denoises the node labels with another mechanism: label denoising with propagation (called Drop in the paper), which also utilizes KProp. The Drop recovers the true labels of nodes based on the fact that nodes with similar labels are more likely to connect with each other. Experimental results show an appropriate privacy-utility trade-off of LPGNN. More info about DP and LDP is presented in subsection \ref{subsec: priv_tech}.

Sajadmanesh et al.~\cite{sajadmanesh2022gap} develop a DP-GNN by adding stochastic noise to the aggregator of GNN such that the existence of a single edge (i.e., edge-level privacy) or a single node and its adjacent edges (i.e., node-level privacy) becomes statistically obscure. Their model consists of three different modules, including an encoder module, an aggregation module, and a classification module. In the encoder module, the private node embedding is learned independent of the edge information. In the aggregation module, they use the graph structure to determine the noisy node embedding after aggregation. In the classification module, a neural network is trained on the private aggregations for node classification without further querying the graph edges. The authors also point out that their model is advantageous to the earlier approaches as it benefits from the multi-hop neighborhood aggregations, and both the edge-level and node-level DP is guaranteed during both the training and the inference stages.

Differentially private stochastic gradient descent (DP-SGD) is a popular algorithm that greatly advances the private deep learning. It can guarantee privacy to all data points in the dataset. Yet, modifications must be made for it to be applicable for graph structured data. This challenge is partially overcame in the work by ~\cite{mueller2022differentially}, where the authors focus on the task of graph-level classification. Furthermore, in this work they attempt to better understand the differences between SGD and DP-SGD through the lens of GNNExplainer\cite{ying2019gnnexplainer}. They find training with DP-SGD learns similar models, but as requiring tighter privacy guarantees (which require higher levels of noise) there is a decline in the similarity to the non-privately trained models. 
Another investigation proposes DP-Adam \cite{daigavane2021node} (i.e., a differentially private Adam-based optimization for GNNs), which achieves similar performance as \cite{mueller2022differentially}.


In addition to the previously mentioned and extensively examined influential works in this field, researchers have pursued similar avenues~\cite{olatunji2021releasing,mueller2022differentially1,lin2022towards,zhang2021graph}, presenting diverse advancements that have resulted in enhanced performance and further stimulating this direction. 



\vspace{-0.5ex}
\subsection{Federated Learning}

Federated Learning (FL)~\cite{konevcny2016federated,yang2019federated, zhang2023towards, zhang2022next} is a promising paradigm to protect data privacy which enables clients (e.g., companies) to train models collaboratively without revealing the raw data. Such a need is prevalent in the real world. For example, several hospitals want to train a model jointly, but their data is not allowed to be shared due to patient privacy concerns. Under this circumstance, FL comes to the rescue via collectively learning models with decentralized data in a privacy-preserving manner. The general framework~\cite{mcmahan2017communication} is to compute local updates from each client, update the global parameters from a central server, and then distribute the update to local clients. In this way, the raw data are only accessible on the local server, preventing information from leakage.

Inspired by the confluence of the development of federated learning and the popularity of graph learning, the interest in the intersection of these two fields, federated graph learning (FGL), has grown rapidly in recent years~\cite{liu2022federated}. Zhang et al.~\cite{zhang2021federated} summarize FGL models into three categories based on how the graph data is distributed, namely Inter-graph FL, Intra-graph FL, and Graph-structured FL.\\




\vspace{-1ex}
\textbf{Inter-graph FL}. 
Inter-graph FL is designed for scenarios in which clients want to jointly train a GNN model, with each of them having a subset of graph samples. 
Every client here has local graph samples. Naturally, the general task for this type of GFL is graph-level prediction (i.e. graph classification). A typical example is drug development, where pharmaceutical companies process confidential datasets, including molecular graphs and their properties. The goal is to utilize the knowledge learned from each organization to build a global GNN model while protecting the raw data. Fedgraphnn~\cite{he2021fedgraphnn} has provided examples of several drug-related tasks.\\

\vspace{-1ex}
\textbf{Intra-graph FL}. 
Intra-graph FL is designed for scenarios in which clients want to jointly train a GNN model, with each of them having a subgraph. Different from inter-graph FL, the graphs in intra-graph are not independent and are related to others. Furthermore, depending on the way how subgraphs are related to each other, it is divided into horizontal and vertical intra-graph FL~\cite{yang2019federated}. As the names suggest, the local sub-graphs are regarded as horizontally/vertically partitioned from the global graph. More specifically, a horizontal partition means that clients share the same feature and label space but different node ID spaces. In contrast, a vertical partition means that clients process different features and label spaces but share the same node ID space. 
For example, in the 
horizontal setting, each user has a local social network from different social networking apps, and the local sub-graphs together form an overall social network, providing extra structural information. For the vertical setting, the users’ properties in different sub-graphs (e.g., local financial graph, local social graph, local knowledge graph) are available, providing different perspectives about the user. Various methods have been designed to protect data privacy in horizontal intra-graph FL~\cite{chen2021fedgl, wufedgnn2021,yao2022fedgcn} and vertical intra-graph FL~\cite{junver2020,mei2019sgnn,ni2021vertical}.\\

\textbf{Decentralized GFL}. 
The above categories mainly study the centralized setting where a central server is required to aggregate local model information from clients as described in the general framework. Although strategies have been applied to prevent data leakage from the local side (e.g., lossy compression before transferring~\cite{zhu2019deep}, noise perturbation~\cite{wei2020federated}), the risk remains in the central server considering it would be possible to infer the protected information. Therefore, it is not practical for clients to accept one of them as the leader (i.e., the central server). From another practical consideration, the existence of a central server becomes a bottleneck due to computation cost and communication overhead. To solve these mentioned issues, 
several decentralized FGL has been developed~\cite{pei2021decentralized, he2021spreadgnn} where clients communicate and aggregate information from each other without a central server.\\

\textbf{Applications}. 
Because of the wide range of applications of graph-federated learning, we provide a separate subsection here. Recommendation, particularly collaborative filtering based on the user-item interaction graph, is one of the most crucial applications of GNN. A federated GNN model is proposed by Huang et al.~\cite{huang2022federated} to ensure personalized recommendations while minimizing the risk of exposure to adversarial attacks. Typically, personalized recommendation requires operations on the entire graph, which could easily lead to the leakage of private user information. To circumvent this, the authors mine the decentralized graph data derived from the global graph. In addition, they introduce a privacy-preserving graph expansion protocol so that high-order information under privacy protection can be incorporated into the model. Consequently, both the local and global information of the graph are incorporated into the model with privacy protection at the cost of little information loss. 
Another important task for GNN is molecular property prediction~\cite{he2021spreadgnn}, which often requires a large amount of training data. However, privacy concerns, regulations, and commercial competition also hinder the collection of a large and centralized training sample. In the work by He et al.~\cite{he2021spreadgnn}, the authors propose SpreadGNN, a decentralized multi-task federated learning GNN model for this task. Notably, SpreadGNN is compatible with partial labels (i.e., missing labels for part of the training sample). In addition, the authors show that SpreadGNN is guaranteed to converge under certain criteria and is effective on multiple datasets with partial labels. Note that one issue associated with using delocalized datasets in federated GNN for molecular property prediction is that the datasets obtained from multiple sources can be highly heterogeneous (i.e. different datasets could cover vastly different regions in the chemical space), which could compromise the model performance, so the authors in ~\cite{zhu2022federated} tackle this issue by proposing federated learning with instance reweighing (FLIT(+)), which is capable of aligning local training across multiple clients. 

Federated GNNs have also been explored in the field of human activity recognition from sensor measurements~\cite{sarkar2021grafehty}. 
Two common major obstacles to human activity recognition are noisy data and privacy. To tackle the two issues, GraFeHty~\cite{sarkar2021grafehty} is proposed, which utilizes a semi-supervised algorithm to tackle the prior issue and the federated learning framework to tackle the later issue. Concretely, in the federated learning framework, only the learned representation is transferred out of the device to the central server so that user privacy can be protected. Also, federated learning allows the authors to address limitations of traditional centralized machine learning for human activity recognition, e.g., infrastructure availability, network connectivity, and latency issues.

\section{Datasets and Applications}

\label{sec: datasets}

This section lists datasets used or could potentially be used to develop privacy attacks and preservation methods for GNN. They can be divided into Social Network, Citation, User-item, Molecule, and Protein. We provide a succinct description for each dataset and their statistics in Table \ref{table:1}. 


\textbf{Social Network}. Social network analysis is a crucial domain that often requires the use of GNN. Social network analysis can be used to detect sub-communities, help marketing, identify disease propagation, and etc. In a social network, nodes are typically users of social media, and edges are the relationships among users. The features of the users in social network typically include gender, education, age, geographical information, relationship status, and etc. 

\begin{itemize}[leftmargin=*]
\item \textbf{Facebook}~\cite{duddu2020quantifying}
Facebook Dataset is a small user-relation network extracted from the Facebook social media. The nodes represent user accounts while the edges describe the connectivity. Each user node has different features including gender, education, hometown, and etc. Leveraging real-world social media data to oppose malicious attacks can be particularly meaningful due to its relevance. 

\item \textbf{Twitter}~\cite{duddu2020quantifying}
Twitter Dataset is a small user-relation network extracted from the Twitter social media. The nodes represent user profiles while the edges describe the connectivity. Node features are user profile information, and the dataset contains information about circles and ego networks. 

\item \textbf{LastFM}~\cite{duddu2020quantifying,he2021node}
LastFM is an Asian social network, where the nodes are the users from Asian countries, and the edges describe the mutual follower relationships. Node features are based on the artists liked by the users. 

\item \textbf{Reddit}~\cite{shokri2017membership, chen2022understanding}
The Reddit dataset is a sub-community of Reddit posts obtained from Sep. 2014. Thus, the nodes are the posts, and an edge exists between two posts if the same user provides comments about both posts. The class label is the community that the post belongs to. 

\item \textbf{Computers}~\cite{chen2022understanding}
Computers is an Amazon co-purchase graph, of which nodes are merchandise, and an edge exists between two merchandises if they are often bought together. The node features are information extracted from the merchandise reviews, and the class labels are the merchandise categories. 

\end{itemize}

\begin{table*}[tbp]
    \footnotesize
    \centering
    \caption{Basic statistics of datasets that have been used or could potentially be used to benchmark privacy attacks and/or preservation methods on GNNs. We collect the sources of all datasets \href{https://github.com/NDS-VU/awesome-gnn-privacy}{here}.}
    \label{table:1}
    \begin{tabular}{ c|c|c|c|c|c|c}
\hline
Dataset Name  & Data Domain    & \# Graphs   &  (Avg.) \# Nodes  & (Avg.) \# Edges  &  \# Features & Ref.   \\ \hline
Facebook  & Social Network    & 1   &  4,039  & 88,234  & - & ~\cite{duddu2020quantifying}  ~\cite{sajadmanesh2020locally}  \\ \hline
Twitter & Social Network & 1 &  81,306  & 1,768,149 & - & ~\cite{duddu2020quantifying}   \\ \hline

LastFM  & Social Network   & 1   &  7,624  & 27,806  &  7,842 & ~\cite{duddu2020quantifying}   ~\cite{he2021node}\\ \hline

Reddit   & Social Network    & 1   &  232,965  & 57,307,946  &  602 & ~\cite{shokri2017membership} ~\cite{chen2022understanding}   \\ \hline

Computers & Social Network  & 1 &  13,752  &  245,861  &  767 & ~\cite{chen2022understanding}   \\ \hline

Cora & Citation   & 1   &  2,708  &  5,429  & 1,433 & ~\cite{duddu2020quantifying} ~\cite{olatunji2021membership}, ~\cite{wu2022model}   \\ \hline

Citeseer  & Citation    & 1   &  3,312  & 4,715  &  3,703 &  ~\cite{duddu2020quantifying}~\cite{olatunji2021membership} ~\cite{wu2022model}  \\ \hline

PubMed  & Citation    & 1  &  19,717   & 44,338  &  500 &  ~\cite{duddu2020quantifying}  ~\cite{wu2022model}  \\ \hline

DBLP  & Citation  & 1   &  17,716   & 105,734  &  1,639 &  ~\cite{shen2022model}  \\ \hline

ogbn-arxiv  & Citation   &  1   &  169,343  & 1,166,243  & 128 &  ~\cite{zhang2020gnnguard}  \\ \hline

Aminer & Citation  & 1   &  659,574  & 2,878,577 &  - & ~\cite{chen2022understanding} \\ \hline

Flixster & User-item   & 1   &  6,000  & 26,173  &  - & ~\cite{ru2021graph}   \\ \hline
Douban & User-item   & 1   &  6,000  & 136,891   &  - &  ~\cite{ru2021graph}   \\ \hline
YahooMusic & User-item   & 1   &  6,000  &  5,335  &  -  & ~\cite{ru2021graph}   \\ \hline

NCI1  & Molecule    &  4,110   &  29.87  & 32.30  &  37 & ~\cite{wu2021adapting, zhang2022inference}  \\ \hline

AIDS  & Molecule   & 2,000   & 15.69 & 16.20  &  42 &  ~\cite{wu2021adapting, zhang2022inference}    \\ \hline

OVCAR-8H & Molecule   & 4,052   & 46.67 & 48.70  &  65 &  ~\cite{wu2021adapting, zhang2022inference}    \\ \hline

PROTEINS & Protein   & 1,113  &  39.06 & 72.82  &  29 &  ~\cite{wu2021adapting}    \\ \hline

ENZYMES & Protein   & 600  &  32.63 & 62.14  &  21 & ~\cite{wu2021adapting}    \\ \hline

    \end{tabular}
    \vspace{-1ex}
    \label{tableapplications}
\end{table*}

\textbf{Citation}. In citation graphs, nodes are papers, and edges characterize the citation relationships among papers. Compared to the social network data, the consequence of adversarial attacks on citation graph is much less serious because the information captured by such dataset is usually not private. However, due to the high accessibility, citation data is still frequently used, and thus we summarize a few below. 

\begin{itemize}[leftmargin=*]
 \item \textbf{Planetoid}~\cite{duddu2020quantifying, olatunji2021membership, wu2022model}
As one most common collection of citation datasets, Planetoid includes Cora, Citeseer and Pubmed, each of which is consisted with scientific publications that are categorized into different classes. The edges describe the citation relationships between the papers. The feature of each node is a 0/1-valued word vector indicating whether a word exists or not. 

 \item \textbf{DBLP}~\cite{shen2022model}
 It is extracted from a website about computer science bibliography. Different from the aforementioned citation graphs, DBLP is heterogeneous as it has four entities including authors, papers, terms, and conferences. 

 \item \textbf{ogbn-arxiv}~\cite{zhang2020gnnguard}
A directed citation graph of which the nodes are computer science arXiv papers indexed by Microsoft academic graph (MAG).

 \item \textbf{Aminer}~\cite{chen2022understanding}
 It consists of multiple relational datasets including citation networks, social networks, and etc. 
\end{itemize}

\textbf{User-item}.
User-item graph is a bipartite graph describing the relationship between users and their interacted items. Leveraging the user-item interaction graph can enable the recommendation based on collaborative filtering. Privacy attack targeting such datasets could cause the information leakage of items liked by certain users, together with the item attributes and user attributes. 

\begin{itemize}[leftmargin=*]
\item \textbf{Flixster}~\cite{ru2021graph}
The rating data of users towards movies. The listed dataset is a small Flixster subset dataset crawled by the authors of ~\cite{ru2021graph}. 

\item \textbf{Douban}~\cite{ru2021graph}
The user-movie interaction graph and connections are useres' comments on the movies. 

\item 
\textbf{YahooMusic}~\cite{ru2021graph}
YahooMusic contains information about the music liked by users. 
\end{itemize}


\textbf{Molecule}.
Graph-based drug discovery has gained increased attention recently due to the development of GNN. Molecules can be represented by graphs, in which nodes are atoms and edges are chemical bonds. Unlike previous datasets in which all data points are used to form one single graph, each data point in molecular dataset is a graph. 

\begin{itemize}[leftmargin=*]

\item 
\textbf{NCI1}~\cite{wu2021adapting, zhang2022inference}
The NCI1 dataset contains molecules that are assessed to be positive and negative to cell lung cancer. In other words, there are only two classes with either 0 or 1 indicating the cancer-target interactivity. 

\item 
\textbf{AIDS}~\cite{zhang2022inference, he2021stealing, zhang2021graphmi}
AIDS dataset is constructed from the AIDS Antiviral Screen Database of Active Compounds, where the molecules are specific to AIDS.

\item 
\textbf{OVCAR-8H}~\cite{wu2021adapting, zhang2022inference}
OVCAR-8H is a database of molecules targeting Ovarian human cancer cell line. 

\end{itemize}

\textbf{Protein}.
Like molecules, proteins can also be represented as graphs with nodes being typically amino acid and edges being the amino acid bond or the spatial proximity. 

\begin{itemize}[leftmargin=*]
\item 
\textbf{PROTEINS}~\cite{wu2021adapting}
It includes two clasess: enzymes or non-enzymes. The nodes are amino acids, and two nodes are connected if they are within 6 Angstroms from each other. 

\item 
\textbf{ENZYMES}~\cite{morris2020tudataset,wu2021adapting}
A dataset containing protein tertiary structures from the BRENDA enzyme database.

\end{itemize}

\vspace{-2ex}
\section{Future directions}\label{sec: future}
While many works have demonstrated the applicability and efficiency of privacy-preserved GNN, newly developed privacy attacks are continuously being introduced that exhibit vulnerabilities in existing privacy-preservation techniques. Additionally, there are generally several avenues yet to be explored, and many challenges to be overcome before reaching a desired level of performance in preserving privacy in real-world applications that are deployed with GNNs. Below we summarize future directions that highlight some of these open challenges and important new frontiers.
\begin{itemize}[leftmargin=*]
\item \textbf{Information leakage for pre-trained GNNs}.
 Pre-training and model-sharing are often used to improve the performance of GNNs under various tasks especially when the labels are inadequate, 
 which has led to an increasing interest in leveraging self-supervised learning for GNNs~\cite{jin2021node,wang2022graph,jin2020self}. 
 Generally, pre-training~\cite{zhou2023comprehensive} can be classified into three categories generally: model-based, mapping-based, and parameter-based methods. The model-based one utilizes the pre-trained source domain as the starting point for the remaining training on the target domain data. This method is also called model-based fine-tuning. The mapping-based approach aligns hidden representations by reducing the difference between the source and target domains. The parameter-based one operates in a multi-task fashion to jointly update a shared network to learn transferable feature representations~\cite{chen2020comprehensive}. However, these strategies are double-edged sword because they can lead to private information leakage and compromise privacy. This issue is particularly concerning since the source and target datasets for the transfer learning are often from different organizations. Thus, it is critical to create a safe environment for different organizations to share the datasets to build the intact transfer learning model while minimizing the privacy concern~\cite{chen2020comprehensive}. Methods have been developed for general deep transfer learning to enhance privacy. For example, Wu et al.~\cite{wu2020characterizing} and Mou et al.~\cite{mou2018generalization} propose stochastic gradient Langevin dynamics (SGLD), yet these for GNNs still await to be developed. 

\item \textbf{GNN in distributed learning settings.} Federated learning with GNN has shown promising results in protecting data privacy, especially in healthcare and financial applications. To reiterate, in FL, clients are able to jointly train a GNN model, with each of them having a sub-graph or subset of graph samples. The focus of most existing works is on the architecture learning and knowledge sharing for building a global GNN model without compromising the privacy of the raw data.
Lyu et al ~\cite{lyu2022privacy} provide a comprehensive survey about the privacy and robustness of federated learning attacks and defenses. In their survey, they cover threat model, privacy attacks and defenses, and poisoning attacks and defenses. However, we want to point out that very few works focus on the attacks on GNN-based federated learning, and more investigation in this area is welcomed. Also, we suggest that it will be beneficial to investigate privacy preservation of GNNs under other distributed learning settings.
 
\item \textbf{Trade-off between privacy and utility.}
As a long-lasting issue for ethical AI, the trade-off between the ethical implication and utility of AI model is almost unavoidable. Metrics for evaluating the defense performance against privacy attack are privacy loss, confidence score, and reconstruction error. The metric for evaluating the model performance is classification accuracy or regression loss. In addition, there lacks a standard way of measuring the trade-off between two groups of metrics, and we believe that studies on these aspects will be particularly helpful. 

\item \textbf{Privacy trade-off in Specialized GNNs}
Specialized GNNs have been developed to mitigate a plethora of data quality challenges, such as imbalanced classification~\cite{ma2023class,wang2021distance, wang2022imbalanced}, mitigating bias~\cite{wang2022improving, wang2022fair, dai2021say}, heterophily~\cite{zheng2022graph,wang2021tree, luan2023graph}, etc.  An investigation of privacy-utility-fairness trade-off in general neural network was done by Marlotte and Giacomo~\cite{pannekoek2021investigating}. In their work, the models under investigation are Simple (S-NN), Fair (F-NN), differentially private (DP-NN), and differentially private and fair neural network (DPF-NN). Similar analysis could be conducted on GNN-based models. Recently, a few works have started to explore this area of privacy and fairness with GNNs~\cite{wang2023individual,wang2023fair}. 
From the imbalance perspective, it would be of interest to study disaggregated performances to better understand which nodes are more susceptible of privacy attacks, and how they might align with the majority/minority groups according to sensitive features and/or class labels. Similarly, such analysis could be done according to node homophily~\cite{mcpherson2001birds}. 


\item \textbf{Privacy in GNNs for Complex Graphs}
While most efforts investigating privacy attacks and preservations in GNNs have focused on simple graphs, in many real-world applications the complex systems are better represented with complex graphs, where dedicated GNN efforts have been made, e.g., on hypergraphs~\cite{feng2019hypergraph,sawhney2021stock}, multi-dimensional graphs~\cite{ma2019multi}, signed graphs~\cite{derr2018signed}, dynamic/temporal graphs~\cite{rossi2020temporal}, knowledge graphs~\cite{schlichtkrull2017modeling}, general heterogeneous graphs~\cite{wang2019heterogeneous,shi2016survey}, etc. It is expected that there could be varying levels of attack/preservation strategies among these complex networks. 

\item \textbf{Generative AI Impacts on Graph/GNN Privacy Attacks/Preservation}
The recent emergence of generative AI in image/NLP domains has raised many privacy concerns, especially in the medical/health domain~\cite{mesko2023imperative}. Also, generated images may include sensitive information that violates companies' copyright and disclose confidential information~\cite{generativeAIproblemIP}. Moreover, the inherent uncertainty in the generation process could even exacerbate the difficulty of designing stable privacy-preserving techniques. Since these generative techniques can be easily adapted to graph-structured data~\cite{fan2023generative}, the same privacy concern may also arise. In molecular generation,  the generated molecules may contain confidential substructures. In the social network domain, if the generation process involves user embeddings, the generated content may reflect the profile information of that user's neighborhood. 

\end{itemize}


\vspace{-2ex}
\section{Conclusion}
\label{sec: conclusion}
In this survey, we present a comprehensive review of the privacy considerations related to graph data and models. We begin by introducing the necessary concepts and notations for understanding the topic of graph privacy. We then provide an overview of various attacks on graph privacy, categorizing them according to the targeted information. We summarize the available techniques for privacy preservation. We also review the datasets and applications that have been used in the study of privacy in graph domains. Finally, we highlight several potential directions for future research in this area. Our hope is that this work will serve as a useful resource for researchers and practitioners interested in this topic, and will encourage further exploration in this promising field.





\vspace{-2ex}
\section*{Acknowledgment}
This research is supported by the National Science Foundation (NSF) under grant number IIS2239881, The Home Depot, and Snap Inc.
This manuscript has been co-authored by UT-Battelle, LLC, under contract DE-AC05-00OR22725 with the US Department of Energy (DOE). The US government retains and the publisher, by accepting the article for publication, acknowledges that the US government retains a nonexclusive, paid-up, irrevocable, worldwide license to publish or reproduce the published form of this manuscript, or allow others to do so, for US government purposes. DOE will provide public access to these results of federally sponsored research in accordance with the DOE Public Access Plan (http://energy.gov/downloads/doe-public-access-plan).

\ifCLASSOPTIONcaptionsoff
  \newpage
\fi



\bibliographystyle{IEEEtran}
\bibliography{references}
%



%

\vspace{-50pt}
\begin{IEEEbiography}
[{\includegraphics[width=1in,height=1.25in,clip,keepaspectratio]{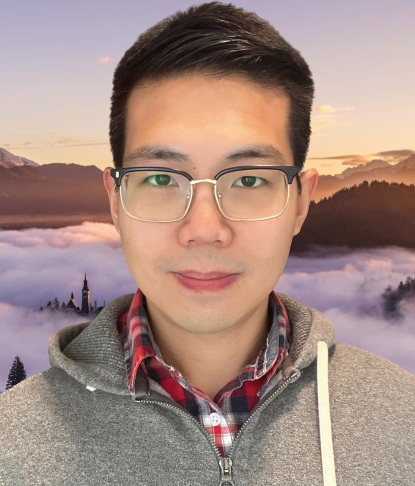}}]{Yi Zhang}  
received his Bachelor degree from the University of Minnesota Twin Cities, where he majored in computer science. He helped complete this work while pursuing a Ph.D. degree in Computer Science at the Vanderbilt University. 
\end{IEEEbiography}

\vspace{-50pt}
\begin{IEEEbiography}
[{\includegraphics[width=1in,height=1.25in,clip,keepaspectratio]{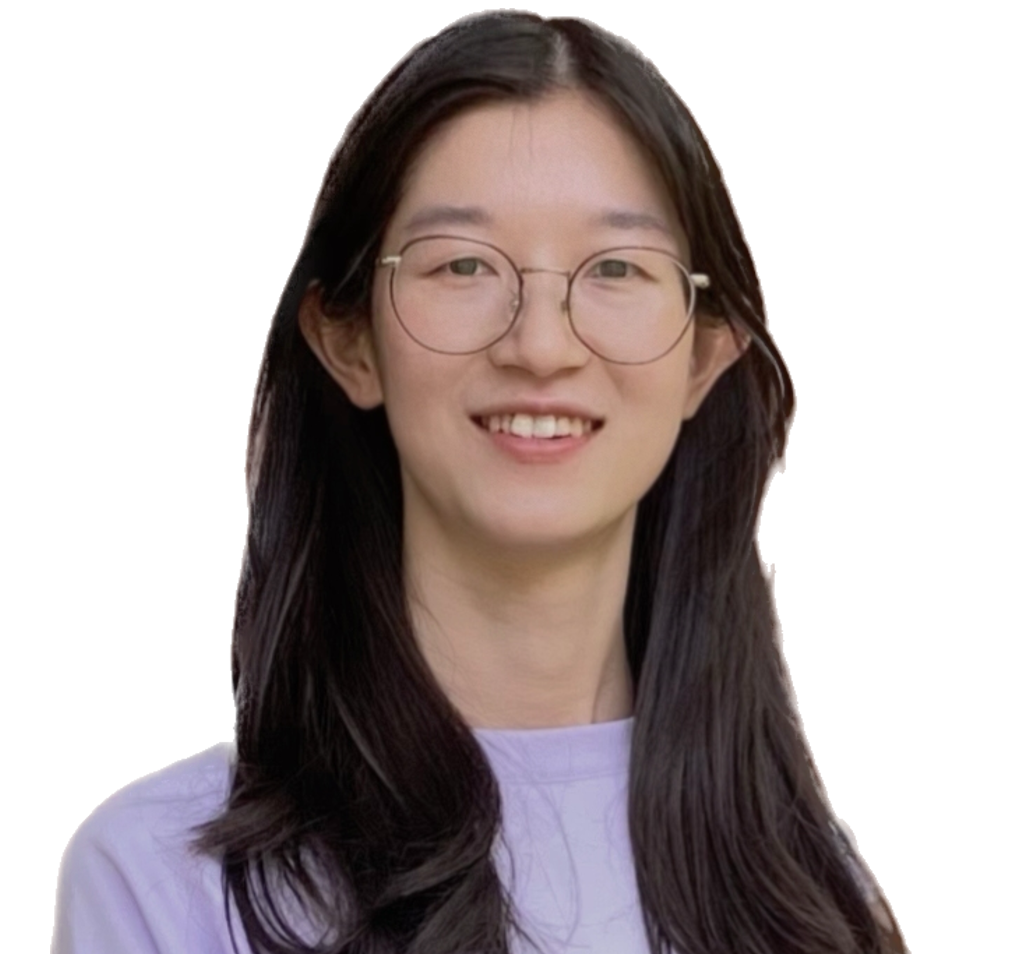}}]{Yuying Zhao} received the B.S. and M.S. degrees in Huazhong University of Science and Technology, China, both with ``Outstanding Graduate” award. She is currently pursuing the Ph.D. degree in Computer Science at Vanderbilt University. Her research interest lies in the intersection of machine learning and graph mining with a special focus on learning beyond-utility perspectives, including fairness, explainability, and diversity. Her papers have been published in AAAI, KDD, WebConf, AMIA, iScience, etc.
\end{IEEEbiography}

\vspace{-50pt}
\begin{IEEEbiography}
[{\includegraphics[width=1in,height=1.25in,clip,keepaspectratio]{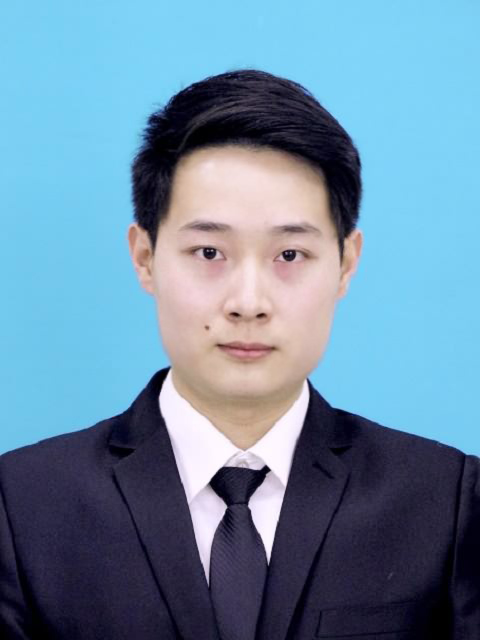}}]{Zhaoqing Li} received the B.S and M.S. degree in automation from Northwestern Polytechnical University, China. He is currently pursuing the Ph.D. degree with the Chinese University of Hong Kong, Hong Kong SAR, China. His current research interests include machine learning and speech recognition.
\end{IEEEbiography}

\vspace{-50pt}
\begin{IEEEbiography}
[{\includegraphics[width=1in,height=1.25in,clip,keepaspectratio]{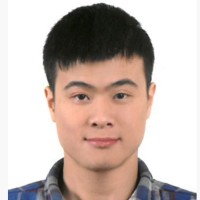}}]{Xueqi Cheng}  
received the B.S. degree from Southwest Jiaotong University, and the M.S. degree from University of Michigan, Ann Arbor. He is currently pursuing the Ph.D. degree in Computer Science Department at Vanderbilt University. His research interests include machine learning and data mining on graphs.
\end{IEEEbiography}

\vspace{-50pt}
\begin{IEEEbiography}
[{\includegraphics[width=1in,height=1.25in,clip,keepaspectratio]{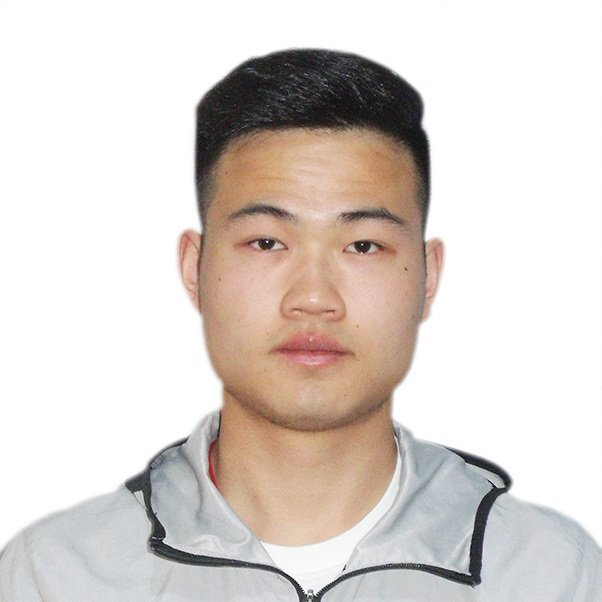}}]{Yu Wang} received the B.S. degree from Harbin Institute of Technology with with 'Outstanding Graduate' award. He is currently pursuing his Ph.D. degree in Computer Science Department at Vanderbilt University. His research interest focuses on machine learning and data mining over graphs with a specific emphasis on overcoming data-quality issues and graph topological learning. He had published papers in KDD, WebConf, CIKM, WSDM, and e.t.c.
\end{IEEEbiography}

\vspace{-50pt}
\begin{IEEEbiography}
[{\includegraphics[width=1in,height=1.25in,clip,keepaspectratio]{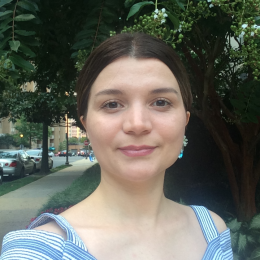}}]{Olivera Kotevska} is a research scientist in Computer Science and Mathematics Division at Oak Ridge National Laboratory (ORNL), Tennessee, USA. She received her Ph.D. degree in Computer Science from the Universite Grenoble Alpes, France in 2018 and during her doctorate she performed research at the National Institute of Standards and Technology (NIST) in Washington, DC. Her research work focuses on privacy algorithms, machine learning, and intersection for various electric grids, human mobility and biomedical applications. She publish and regularly serves as a program committee member at the top conferences in these domains and served in organizational roles including Co-Chair of Privacy Algorithms for Systems workshop, Co-Chair of IEEE Big Data Industry and Government Program, Chair of IEEE Power and Energy Society Computational Analytical Methods Subcommittee, and Co-Editor of Sensors journal special issue IoT Data Analytics. Dr.Kotevska is an organizer and chair of IEEE WiE East Tennessee affinity group. She received IEEE Senior membership award in ’21 and ORNL CSMD Outreach and Service award in ’20. 
\end{IEEEbiography}

\vspace{-50pt}
\begin{IEEEbiography}
[{\includegraphics[width=1in,height=1.25in,clip,keepaspectratio]{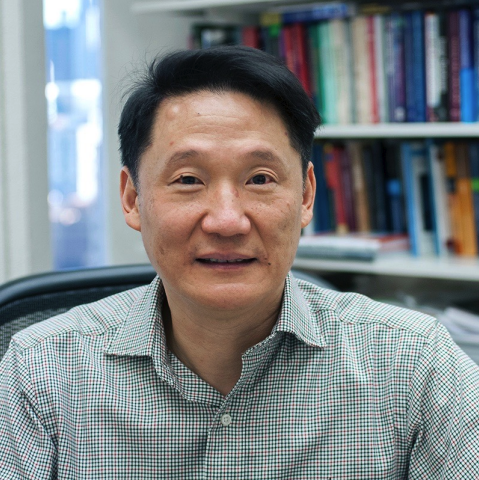}}]{Philip S. Yu}  
received the B.S. Degree in E.E. from National Taiwan University, the M.S. and
Ph.D. degrees in E.E. from Stanford University, and the M.B.A. degree from New York University. He is a Distinguished Professor in Computer Science at the University of Illinois at Chicago and also holds the Wexler Chair in Information Technology. Before joining UIC, Dr. Yu was with IBM, where he was manager of the Software Tools and Techniques department at the Watson Research Center. His research interest is on big data, including data mining, data stream, database and privacy. He has published more than 1,200 papers in refereed journals and conferences. He holds or has applied for more than 300 US patents. Dr. Yu is a Fellow of the ACM and the IEEE. Dr. Yu is the recipient of ACM SIGKDD 2016. Innovation Award for his influential research and scientific contributions on mining, fusion and anonymization of big data, the IEEE Computer Society’s 2013 Technical Achievement Award for “pioneering and fundamentally innovative contributions to the scalable indexing, querying, searching, mining and anonymization of big data”, and the Research Contributions Award from IEEE Intl. Conference on Data Mining (ICDM) in 2003 for his pioneering contributions to the field of data mining. He also received the ICDM 2013 10-year Highest-Impact Paper Award, and
the EDBT Test of Time Award (2014). He was the Editor-in-Chiefs of ACM Transactions on Knowledge Discovery from Data (2011-2017) and IEEE Transactions on Knowledge and Data Engineering (2001-2004). 
\end{IEEEbiography}

\vspace{-44pt}
\begin{IEEEbiography}
[{\includegraphics[width=1in,height=1.25in,clip,keepaspectratio]{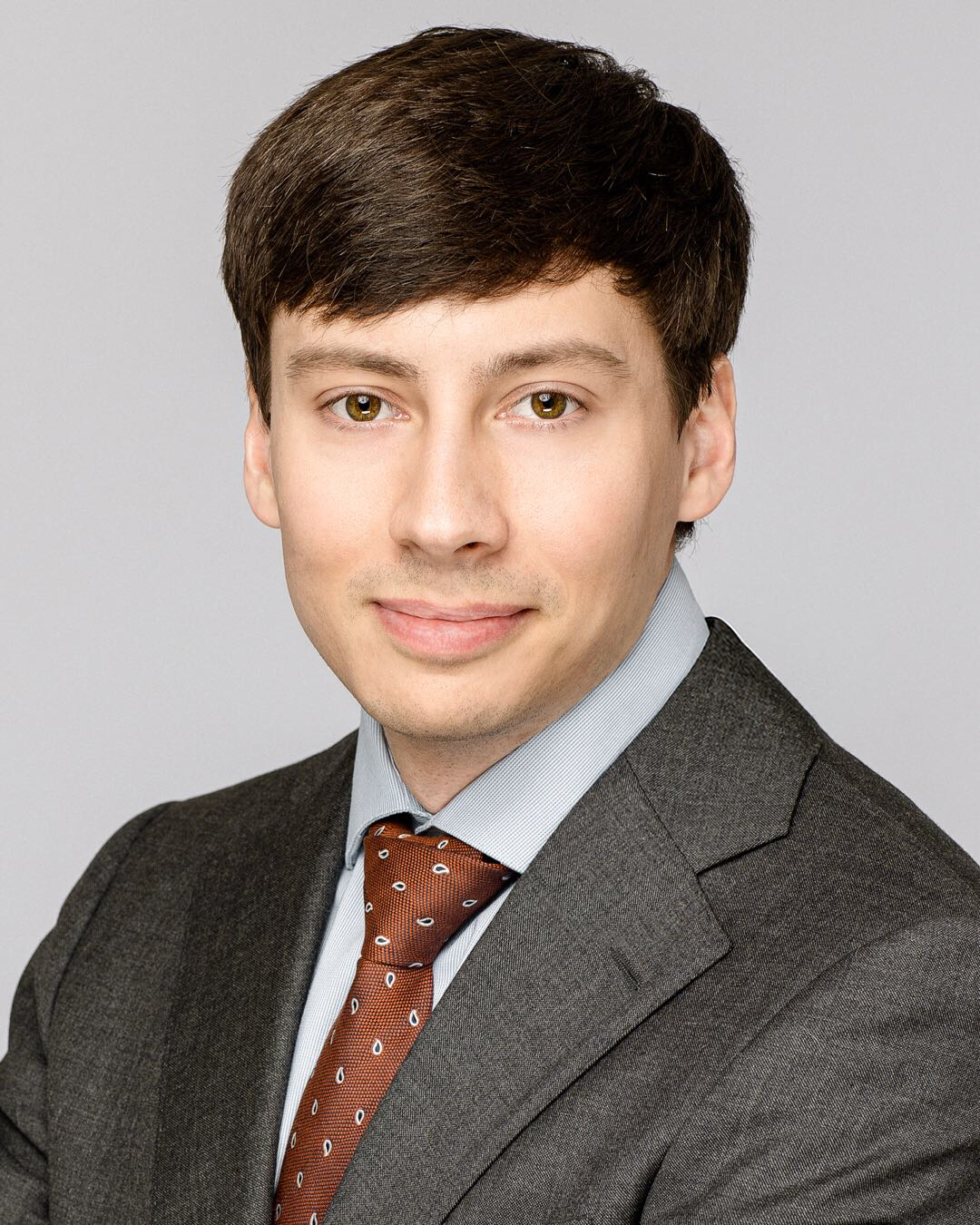}}]{Tyler Derr}  
is an Assistant Professor at Vanderbilt University in the Department of Computer Science and Data Science Institute. He earned his PhD in Computer Science from Michigan State University in 2020. His research focuses on data mining and machine learning, with emphasis on social network analysis, deep learning on graphs, and data science for social good. He is actively involved in top conferences in his field, both in terms of publishing and serving as an SPC/PC member, while receiving recognition such as the Best Student Poster Award at SDM'19 and Best Reviewer Awards at ICWSM'19 and '21, as well as WSDM'22. He has contributed to the organization of international conferences, including serving as the Publicity Co-Chair of KDD'22 and '23, Doctoral Consortium Co-Chair of WSDM'22, and Proceedings Co-Chair of KDD'21. Being passionate about sharing knowledge, he has co-organized multiple workshops including Machine Learning on Graphs (MLoG) Workshop at WSDM’22 and ’23 along with at ICDM’22 and ’23; he has delivered tutorials on Graph Neural Networks at KDD’20 and AAAI’21. Additionally, he was honored with the Fall 2020 Teaching Innovation Award from the School of Engineering at Vanderbilt University, highlighting his dedication to exceptional teaching. Tyler received the NSF CAREER Award in 2023. For more detailed information, please visit his website at https://www.TylerDerr.com. 
\end{IEEEbiography}








\end{document}